\documentclass[journal]{new-aiaa}
\usepackage[utf8]{inputenc}
\usepackage{textcomp}

\usepackage{graphicx, multirow}
\usepackage{caption}
\usepackage{subcaption}
\usepackage{amsmath}
\usepackage[version=4]{mhchem}
\usepackage{siunitx}
\usepackage{gensymb}
\usepackage{longtable,tabularx}
\usepackage{nicefrac}
\title{Multidisciplinary Design Optimization of Reusable Launch Vehicles for Different Propellants and Objectives}

\author{K. Dresia \footnote{Ph.D. Candidate, Institute of Space Propulsion, kai.dresia@dlr.de},
S. Jentzsch \footnote{M.Sc. Candidate, RWTH Aachen University, simon.jentzsch@rwth-aachen.de},
G. Waxenegger-Wilfing \footnote{Research Scientist, Institute of Space Propulsion, guenther.waxenegger@dlr.de},
R. Hahn \footnote{Research Scientist, Institute of Space Propulsion, robson.dossantoshahn@dlr.de} and
J. Deeken \footnote{Head of System Analysis Group, Institute of Space Propulsion, jan.deeken@dlr.de}}
\affil{Institute of Space Propulsion, DLR, Hardthausen, Germany}

\author{M. Oschwald \footnote{Head of Rocket Propulsion Department, Institute of Space Propulsion, michael.oschwald@dlr.de}}
\affil{Institute of Space Propulsion, DLR, Hardthausen, Germany}
\affil{Institute of Jet Propulsion and Turbomachinery, RWTH Aachen University, Aachen, Germany}

\author{F. Mota \footnote{Professor, Aerospace Engineering Department, mota.fabio@ufabc.edu.br}}
\affil{Federal University of ABC, São Paulo, Brazil}

\begin{document}

\maketitle
\onehalfspacing

\begin{abstract}
Identifying the optimal design of a new launch vehicle is most important since design decisions made in the early development phase limit the vehicles' later performance and determines the associated costs. Reusing the first stage via retro-propulsive landing increases the complexity even more. Therefore, we develop an optimization framework for partially reusable launch vehicles, which enables multidisciplinary design studies. The framework contains suitable mass estimates of all essential subsystems and a routine to calculate the needed propellant for the ascent and landing maneuvers. For design optimization, the framework can be coupled with a genetic algorithm. The overall goal is to reveal the implications of different propellant combinations and objective functions on the launcher's optimal design for various mission scenarios. The results show that the optimization objective influences the most suitable propellant choice and the overall launcher design, concerning staging, weight, size, and rocket engine parameters. In terms of gross lift-off weight, liquid hydrogen seems to be favorable. When optimizing for a minimum structural mass or an expandable structural mass, hydrocarbon-based solutions show better results. Finally, launch vehicles using a hydrocarbon fuel in the first stage and liquid hydrogen in the upper stage are an appealing alternative, combining both fuels' benefits.

\end{abstract}
 \newpage
\section*{Nomenclature}
{\renewcommand\arraystretch{1.0}
\noindent\begin{longtable*}{@{}l @{\quad=\quad} l@{}}
$\Delta v$ & delta-v [\si[per-mode=symbol]{\meter \per \second}] \\
$\varepsilon$   & structural coefficient [-] \\
$g_0$   &   standard gravity [\si[per-mode=symbol]{\meter \per \second \squared}] \\
$m$     & mass [\si{\kilo \gram}] \\
$r$ & stage radius [\si{\meter}] \\
ROF & mixture ratio [-] \\
$t_b$ & burn time [\si{\second}] \\ [1ex]
\multicolumn{2}{@{}l}{Subscripts}\\
1 & first stage\\
2 & second stage\\
c   & combustion chamber \\
i   & stage number  \\
p   & propellant \\
pl  & payload \\
s   & structure \\
sl  & sea level \\
vac & vacuum \\ 

\end{longtable*}}

\section{Introduction}
\lettrine{T}{he} early development phase of a new launch vehicle is critical because the selected concepts fix the majority of the expected costs \cite{blair2001}. But the early development phase is also challenging, as many subsystems of the vehicle are closely interlinked and influence each other's design choices. For the optimal global design, multidisciplinary design studies are necessary that consider all mission constraints and design variables at the same time. By balancing different disciplines, such as structure, propulsion, aerodynamics, or economic factors, multidisciplinary design studies can not only increase the performance of a design concept but also reduce its expected costs. 

Multidisciplinary design studies have long been used to design space launch vehicles. They have always focused on the most promising, state-of-the-art technology for a powerful and cost-effective launch vehicle. During the Space Shuttle era, studies aimed to optimize a fully reusable, winged launch vehicle because this concept was thought to be the most economical \cite{braun1996, rowell1996, koelle2000}. After the idea of full reusability was discarded, the focus shifted to expendable launchers \cite{durante2004, bayley2007, briggs2007}, demonstrating that multidisciplinary design optimization could lower the expected costs. In his study, Castellini et al. \cite{castellini2010, castellini2012} developed a multidisciplinary design optimization framework for expendable launch vehicles. The framework includes well-known empirical and well-validated mass and performance estimates for liquid rocket engines and the main structural and non-structural components of launch vehicles (e.g., propellant tanks, payload adapters, and fairings). 

For some years now, economic and technical considerations have complicated the design of new launch vehicles even further: With the success of SpaceX's partially reusable Falcon 9, the international market of launch services gained a promising competitor to the traditional expendable launch vehicles. Reusing the rocket's first stage after landing it through retro propulsion allows SpaceX to offer a significantly lower price than other launch service providers; thus, reducing costs became the driving design factor for future launch vehicles. Furthermore, with this paradigm shift from high-performance expendable launch vehicles to low-cost reusable rockets, propellants' choice also needed to be reevaluated. Cryogenic methane is of great interest for future launch vehicles \cite{patureaudemirand2019, ukai2019, kajon2019, traudt2019} and its properties are intensively researched \cite{soller2014, borner2018, haemisch2019, waxenegger-wilfing2020}. Cryogenic methane offers a potentially more economical alternative to cryogenic liquid hydrogen and kerosene. Because no operational launch vehicle using methane has been built, its impact on the overall launch vehicle configuration needs to be assessed and compared with traditional configurations. Finally, a vertical landing reusable first stage increases the complexity and the couplings between the launcher's subsystems even more. Additional components (e.g., the landing gear) and new mission phases (e.g., the landing maneuver) need to be considered during the design phase of a new launch vehicle. 

The objective function used in a multidisciplinary design study largely influences the study's outcome. Earlier studies (compare the work of Balesdent \cite{balesdent2012} for a comprehensive overview) tried to minimize the gross lift-off weight, the structural mass, or the expected total costs, which were determined by empirical cost models \cite{wertz2004, koelle2013}. Ideally, one would always want to optimize a new launch vehicle for its expected costs, but detailed cost models, especially for new technologies, are prone to errors. For (partially) reusable launch vehicles, simple mass-based objective functions might lead to non-optimal configurations because they might not directly correlate with the expected costs. So far, however, the literature lacks a discussion about the impact of different objective functions for reusable launch configurations. 

In order to meet the new design challenges of future launch vehicles, various multidisciplinary design studies have recently been conducted. Balesdent et al. \cite{balesdent2016} studied, among other things, the influence of technological uncertainties (e.g., new propellant combinations or reusable rocket engines) on the design process of future launch vehicles. Briese et al. \cite{briese2020} presented a modular multidisciplinary modeling framework for reusable launch vehicles with a particular focus on the trajectory optimization of the ascent and descent phases. On a component level, Vietze et al. \cite{vietze2018} developed a toolbox to optimize the structure, geometry, and thermal protection system of a cryogenic launcher stage. Stappert et al. \cite{stappert2019} reviewed different return methods and propellant combinations for a reusable first stage, taking into account preliminary design assumptions. Moroz et al. \cite{moroz2019} even applied artificial intelligence to automatically consider reusability aspects during the preliminary design phase, such as the expected life of the turbopumps. Brevault et al. \cite{brevault2020} optimized a multi-mission launcher family with a winged first stage and cryogenic methane as fuel in both stages.

Although some research has been carried out on multidisciplinary design optimizations for reusable launchers, no studies have been found that simultaneously include retro-propulsive landing, the detailed modeling of the rocket engine, and a comparison of different propellant combinations and objective functions. However, this investigation is essential to find the optimal architecture for future, cost-effective reusable launch vehicles.


This paper aims to optimize a two-stage, partially reusable launch vehicle without making preliminary assumptions on the propellant choice, stage separation velocity, general launcher architecture (e.g., number of engines), or internal engine parameters (e.g., combustion chamber pressure). We examine different propellant choices for each stage (LOX/LH2, LOX/RP-1, and LOX/CH4) and objective functions (gross lift-off weight, structural mass, and a newly defined expendable structural mass). The reusable first stage lands via retro-propulsion, similar to SpaceX's Falcon 9 (downrange landing). To investigate the influence of uncertainties during pre-development (e.g., uncertainties due to new technologies), we perform a sensitivity analysis of the engine performance and mission delta-v budget. The overall goal is to reveal the implications of different propellant combinations and objective functions on the launcher's optimal design for various mission scenarios. As our main contribution, we

\begin{itemize}
  \item develop an optimization framework for a reusable launch vehicle;
  \item include structure and propellant mass estimates for reusability aspects (e.g., landing gear, landing maneuvers);
  \item apply genetic algorithms for launch vehicle optimization;
  \item review the implications of different propellant combinations and objective functions on the design of the launcher and its engines. In this work, the focus is on gas-generator cycle engines.
\end{itemize}

The remainder of this paper is structured as follows: Chapter~\ref{s:tool} describes the optimization framework for a partially reusable launch vehicle. The framework contains suitable mass and performance estimates for each subsystem of the launch vehicle, including the liquid rocket engines. Additionally, the functional principle of genetic algorithms is explained. Chapter~\ref{s:results} presents the mission scenario, assumptions, and results of the optimization algorithm. Chapter~\ref{s:discussion} discusses the insights for the design of future reusable launch vehicles. Finally, Chapter~\ref{s:conclusion} provides concluding remarks.

\section{Reusable Launch Vehicle Optimization Tool}
\label{s:tool}
The reusable launch vehicle optimization tool is a framework that can design a new, virtual launch vehicle based on high-level mission parameters, such as payload mass and delta-v requirements. The framework assembles an entire rocket based on these requirements and estimates the characteristics of all internal subsystems (e.g., propellant tanks) according to the expected loads. In general, the tool can choose from different propellant combinations, a variable number of stages, and multiple engines per stage. The first stage of the launch vehicle might be reusable and might land via retro-propulsive landing. A general and detailed description of the underlying mass- and performance calculations are given in two preceding theses~\cite{christall2017, jentzsch2020}. 

The first part of the following section describes the general program flow with a particular focus on a two-stage, partially reusable launch vehicle. The second part deals with finding the optimum launch vehicle configuration for given mission requirements and constraints by using a genetic algorithm. The same approach was already used to study different engine cycles~\cite{waxenegger-wilfing2018}.

\subsection{Launch Vehicle Model}

\begin{figure}[h!]
\centering
\includegraphics[width=0.7\linewidth]{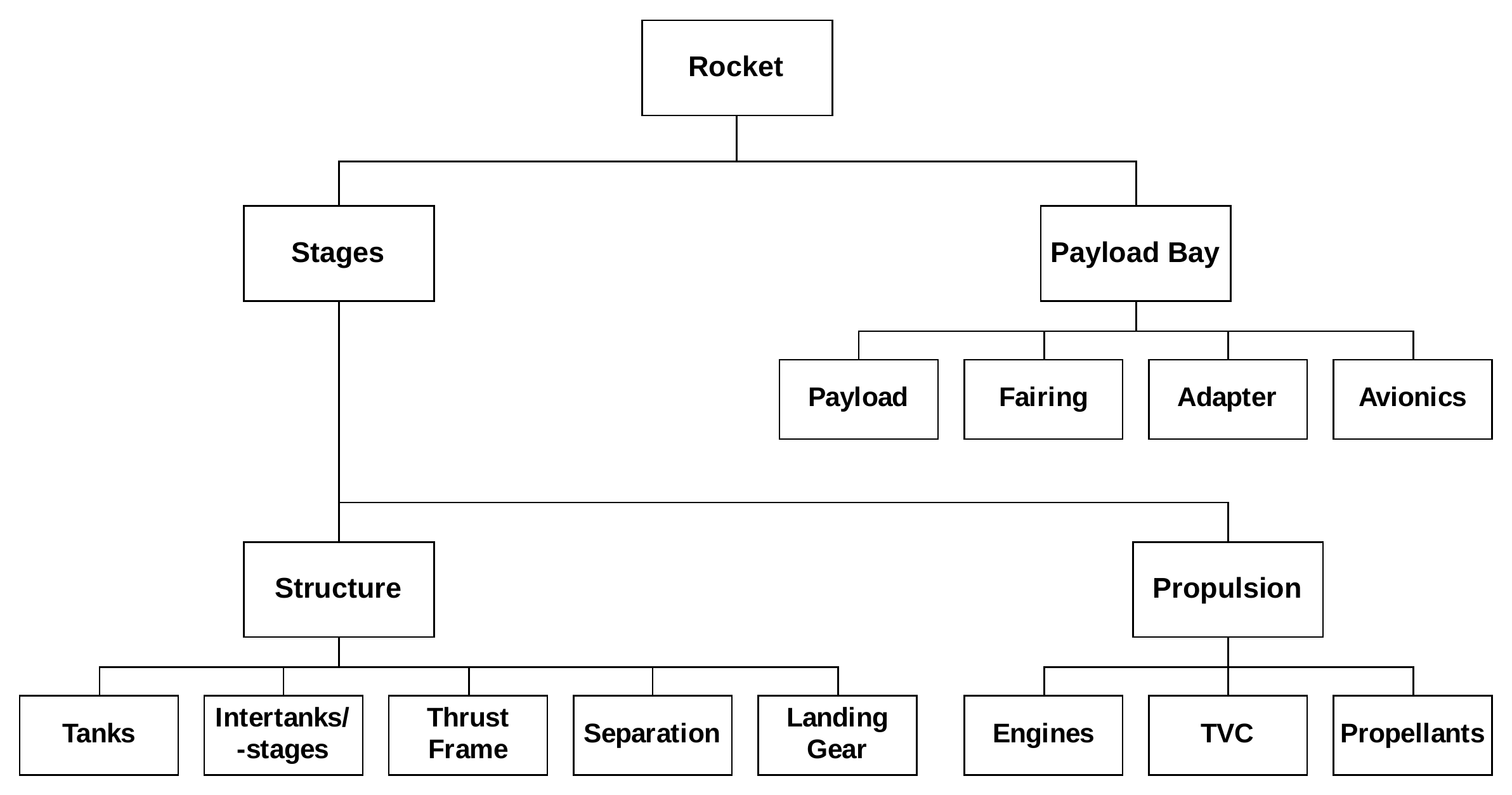}
\caption{Launch Vehicle Composition}\label{fig:fig2.1}
\end{figure}

A launch vehicle consists of its stages and the payload bay, which accommodates payload, payload adapter, avionics, and fairing. Each stage consists of its structure, including the propellant tanks, intertank, interstage, thrust frame, the means to separate itself from the upper stage or payload, its engine(s), the thrust vector control system, and the propellants. Fig.~\ref{fig:fig2.1} illustrates the composition of the launch vehicle.

For essential subsystems, mass estimations are implemented according to Castellini~\cite{castellini2012}. The propellant tanks are modeled as separate cylindrical tanks with spherical lids and calculated according to Barlow's formula with a safety factor of 1.5. We further consider additional stringer and ring frame reinforcements as well as insulation if necessary~\cite{jentzsch2020}. To consider the mass of the landing gear, e.g., landing legs and grid fins, the first stage dry mass is increased by \SI{15}{\percent}. This weight penalty for the landing gear was estimated by reverse-engineering SpaceX's Falcon 9. To consider uncertainties in the mass estimates, neglected components, and neglected additional loads (e.g., due to gusts), additional mass margins are also applied. The margins are set to \SI{10}{\percent} for the upper stage's and \SI{15}{\percent} for the first stage's dry mass. Similar to Stappert et al. ~\cite{stappert2018}, a higher margin is applied to the first stage to reflect more substantial uncertainties about the landing gear structures and additional loads.

\subsection{Virtual Launch Vehicle Assembly}
Assembling a new virtual launch vehicle that meets all mission requirements and constraints is an iterative process. First, the launcher is built with a predefined number of engines in each stage. High-level engine performance characteristics ($I_{sp}$, thrust, and mass) are estimated with a detailed engine model based on the selected propellants, mixture ratios, chamber pressure, and the engine's geometry. Secondly, the launcher's entire structure is built, and the required propellant masses for the first and upper stages are calculated. Finally, the thrust-to-weight ratio and the minimum acceleration of each stage of the launch vehicle is checked against user-defined constraints. If the thrust-to-weight ratio of a specific stage is too small, an additional engine is added to this stage, and the design of the launch vehicle starts from the beginning. The following sections illuminate the rocket engine model and explain the propellant mass calculations in more detail, focusing on a partially reusable launch vehicle.

\subsubsection{Propulsion System}
The engine characteristics are estimated with NASA's CEA program~\cite{CEAmanual}. It can be used to obtain chemical equilibrium compositions of complex mixtures, and it can deliver theoretical rocket performance parameters. As the accuracy of the generated data varies with the combustion chamber pressure, several currently operational or historical engines were recalculated, and a correction formula was implemented based on a regression analysis of the $I_{sp}$ deviations, similar to Castellini~\cite{castellini2012}. For the assumed gas-generator engines, a maximum turbine pressure ratio of 20, turbine and pump efficiencies of \SI{50}{\percent}, and a gas-generator temperature of \SI{900}{\kelvin} are assumed. These values are used to estimate the additional propellant mass flow and the performance losses of the open engine cycle. 


\subsubsection{Propellant Mass Calculations}
A reusable first stage that lands downrange through retro-propulsion cannot expend all of its propellants during ascent. The propellant amount that needs to be reserved for reentry and the landing burn depends on the size and weight of the first stage. Therefore, traditional staging optimization techniques for expendable launchers are not applicable, and a different approach is necessary.

The propellant mass calculations for each stage, including the propellant mass needed for the landing of the first stage are presented below. The calculations require the following input values: the payload bay mass, the fairing mass, the specific impulse of the engines, an initial values of the structural coefficients for both stages, the total delta-v of the mission as well as its allocation between the first stage ($\Delta v_\text{1,ascent}$) and the upper stage ($\Delta v_\text{2}$). Furthermore, the delta-v equivalent for the reentry and landing burn ($\Delta v_\text{1,landing}$) is required. The formulas for the structural coefficient $\varepsilon$, the mass ratio $\nicefrac{m_\text{0}}{m_\text{f}}$, and the Tsiolkovsky equation are the basis of the derived formulas for the propellant mass calculations:
\begin{equation}
\varepsilon_{i}=\frac{m_\text{s,i}}{m_\text{s,i}+m_\text{p,i}} \qquad
\left( \frac{m_\text{0}}{m_\text{f}} \right)_{\text{i}} =\frac{m_\text{s,i}+m_\text{p,i}+m_\text{pl,i}}{m_\text{s,i}+m_\text{pl,i}} \qquad
\Delta v_\text{i}=g_{0} \cdot I_{sp,\text{i}} \cdot ln \left(\frac{m_\text{0}}{m_\text{f}} \right)_\text{i} \qquad
\label{eq:eq2.1}
\end{equation}

Here, $m_\text{s,i}$ is the structural mass, $m_\text{p,i}$ the propellant mass, and $m_\text{pl,i}$ the payload mass of a particular stage $i$. $m_\text{0}$ denotes the initial and $m_\text{f}$ the final mass of the stage. $g_0$ is the standard gravity. Now, the propellant mass of the upper stage can be calculated using the payload bay mass $m_\text{pl}$, the fairing mass $m_\text{fairing}$, as well as the delta-v $\Delta v_\text{2}$, the vacuum specific impulse $I_{sp\_\text{vac,2}}$ and the structural coefficient $\varepsilon_\text{2}$ of the upper stage:
\begin{equation}
m_\text{p,2}=\left(m_\text{pl}-m_\text{fairing}\right)\frac{1-\exp \left(\frac{\Delta v_\text{2}}{I_{sp\_\text{vac,2}} \cdot g_{0} }\right)}{\frac{1}{\varepsilon_\text{2}-1} \left( 1-\varepsilon_\text{2} \exp \left(\frac{\Delta v_\text{2}}{I_{sp\_\text{vac,2}}\cdot g_{0}}\right) \right)}  \qquad
\label{eq:eq2.2}
\end{equation}
Furthermore, it is necessary to obtain an expression for the first stage structural coefficient after stage separation $\varepsilon_\text{1,landing}$ to calculate the first stage propellant mass and its division into propellant mass for the ascent and the descent. The structural coefficient is determined with the delta-v delivered by the engines during descent $\Delta v_\text{1,landing}$ and the mean specific impulse of the first stage engines $I_{sp,1}$. $\Delta v_\text{1,landing}$ is the sum of the delta-v for the reentry burn and the landing maneuvers:
\begin{equation}
\varepsilon_\text{1,landing}=\frac{m_\text{s,1}}{m_\text{s,1}+m_\text{p,1,landing}}= \left( \frac{m_\text{0}}{m_\text{f}} \right)_\text{landing}^{-1} = -\exp \left(\frac{\Delta v_\text{1,landing}}{ I_{sp,\text{1}} \cdot g_{0} } \right)
\qquad
\label{eq:eq2.3}
\end{equation}

Reverse engineering of SpaceX's Falcon 9 rocket and the application of a \SI{5}{\percent} margin led to the assumption of $\Delta v_\text{1,landing}=2000~\nicefrac{m}{s}$ for $\Delta v_\text{1,ascent}=3500~\nicefrac{m}{s}$. For potential launch vehicles with a greater or smaller first stage ascent delta-v, $\Delta v_\text{1,landing}$ must be adapted accordingly. It is assumed that a higher/lower velocity at stage separation entails a longer/shorter reentry burn. Thus, the first stage's velocity upon entering the denser parts of the atmosphere after the reentry burn is similar for all configurations. The deceleration due to drag in the atmosphere's denser parts can be approximated with the terminal velocity:

\begin{equation}
v_\text{1, terminal}= \sqrt{\frac{2mg}{\rho A C_d}}
\label{eq:terminal}
\end{equation}

Assuming a constant drag coefficient $C_d$ for all rockets, the terminal velocity scales with the square root of the rocket's projected area and its mass after the reentry burn. Table \ref{tab:ballistic} shows that this coefficient is quite similar for different fuels and first stage sizes of the optimized rockets. Therefore, we can assume a similar deceleration behavior due to atmospheric drag, leading to similar delta-v requirements for the landing burn.

\renewcommand{\arraystretch}{1.6}
\begin{table}[h]
        \centering
        \begin{tabular}{l c c}
            \hline\hline
                  \textbf{Prop.} & \textbf{$\Delta v_\text{1,ascent}$} [\si{\kilo \metre \per \second}] & \textbf{$\sqrt{\nicefrac{m}{A}}$ [$\si{\tonne}^{\nicefrac{1}{2}} \si{\meter}^{-1}$]}\\
            \hline
           LH2                    & 2.9 / 3.0 / 4.3  & 1.49 / 1.48 / 1.56    \\
           LCH4                 & 2.7 / 3.1 / 4.5  & 1.45 / 1.46 / 1.57    \\
           RP-1                    & 3.0 / 3.2 / 4.4  & 1.48 / 1.47 / 1.62    \\
 \hline \hline
	\end{tabular}
	\caption{Ratio of Mass and Projected Area for the GLOW-, SM-, and EM-Optimized Rockets in Table \ref{tab:tab4.1}}
	\label{tab:ballistic}
\end{table}

\noindent Finally, the first stage structural mass can be calculated: 
\begin{equation}
m_\text{s,1}=m_\text{0,2}\frac{1-\exp \left(\frac{\Delta v_\text{1,ascent}}{I_{sp,1}\cdot g_{0}}\right)}{\frac{1}{\varepsilon_\text{1,landing}} \exp \left(\frac{\Delta v_\text{1,ascent}}{I_{sp,1}\cdot g_{0}}\right)-\frac{1}{\varepsilon_{1}}} \qquad
\label{eq:eq2.4}
\end{equation}
$m_\text{0,2}$ is the total upper stage mass before engine ignition (including the payload bay), $\Delta v_\text{1,ascent}$ the first stage ascent delta-v, $I_{sp,1}$ the mean specific impulse of the first stage engines during ascent, $\varepsilon_{1}$ the first stage structural coefficient and $\varepsilon_\text{1,landing}$ the previously calculated first stage structural coefficient after stage separation. Then, the total first stage propellant mass, the propellant mass for first stage landing, and the first stage ascent propellant mass can be derived:
\begin{equation}
m_{p,1}=m_\text{s,1} \frac{1-\varepsilon_\text{1}}{\varepsilon_\text{1}} \qquad 
m_\text{p,1,landing}=m_\text{s,1} \frac{1-\varepsilon_\text{1,landing}}{\varepsilon_\text{1,landing}} \qquad
m_\text{p,1,ascent}=m_\text{p,1}-m_\text{p,1,landing} \qquad
\label{eq:eq2.5}
\end{equation}

\subsubsection{Iteration and Convergence}
As the structural coefficients of the first and upper stages depend on their structural masses, which are unknown at the time of propellant mass calculation, initial values need to be chosen for the first iteration. After the propellant mass of the upper stage is determined, its structural mass can now be calculated using the mass estimations, and a new structural coefficient is derived for the next iteration. This process is repeated until the convergence of the upper stage structural coefficient. Subsequently, the same procedure is carried out for the first stage. 

For the first stage, the mean $I_{sp}$ represents an additional convergence criterion. The mean $I_{sp}$ of the first stage during ascent is unknown in advance because it depends on the ascent trajectory and the particular rocket engines. For convergence, the mean $I_{sp}$ is derived via a stepwise engine performance parameter calculation in a 2D-trajectory simulation until main engine cut-off. The gravity turn is simulated with a start altitude of \SI{250}{\metre}, a final pitch angle of \ang{25} and a constant turn rate of \SI[per-mode=symbol]{0.45}{\degree\per\second}. With the implementation of an atmospheric model, the rockets position, velocity, and acceleration are determined in a two-dimensional space with a time step of \SI{1}{\second} until the first stage propellant mass for the ascent is spent and the engines are cut off. This is necessary to obtain an accurate value for the mean $I_{sp}$, as the ambient pressure changes during the ascent, thus, influencing the expansion in the nozzle.

\subsection{Validation with existing Launch Vehicle}
To validate the model of the two-stage launch vehicle with a reusable first stage, we use the Falcon 9 v1.2 Block 5 rocket as a reference vehicle because it is currently the only operational launch vehicle with this configuration. Its first stage is powered by nine Merlin engines and the upper stage by one Merlin engine featuring a larger nozzle with a higher expansion ratio. The gas-generator engines burn a LOX/RP-1 propellant mixture. 

As SpaceX is a private company, not much official information is available and technical data is published very scarcely. Furthermore, the Falcon 9 rocket's rapid evolution resulted in many modifications; thus, various versions of the vehicle, making a distinction difficult. Therefore, the data used hereafter is based on unofficial estimations and cannot be expected to be \SI{100}{\percent} precise. However, the values are assumed to describe Falcon 9 sufficiently accurately.

We use the subroutines of the virtual launch vehicle assembly program to calculate the characteristics of a system similar to the Falcon 9. The goal is to validate the mass estimates and the propellant calculations. We choose a \SI{5000}{\kilo \gram} payload GTO mission, which has already been demonstrated by Falcon 9 in the reusable configuration. We further fix the delta-v contributions and the number of engines in the first and upper stage. Other design parameters are chosen to match the real-world Falcon 9 as closely as possible. 

Table \ref{tab:tabA.1} in the appendix shows that we can produce a launch vehicle that is strikingly similar to the real Falcon 9 with only minor deviations. Our simplified propellant tank models result in slightly larger tanks that contain more propellant. Due to the increased weight, an engine with a slightly higher thrust has to be selected. Other high-level parameters like the engine's specific impulse at sea level and in vacuum match the reference values very well.

\subsection{Genetic Algorithm}

The previous sections dealt with the reusable launch vehicle framework, which can design new, virtual launch systems based on high-level mission requirements and design parameters. This section describes how to find the optimum set of design parameters for a given optimization goal, referred to as the objective function. Genetic algorithms, a class of evolutionary algorithms, are numerical optimization algorithms inspired by natural selection and natural genetics~\cite{coley1999}. Biological evolution, first formulated by Charles Darwin, builds the basis of this optimization method, which mimics the adaptive change of species through natural selection, reproduction, and the occurrence of mutations in light of the current environment \cite{back1996}. Typically, a genetic algorithm uses the components~\cite{coley1999}:
\begin{itemize}
\item A randomly generated population, representing several guesses of the solution to the problem
\item A method for evaluating the quality of the individual solutions within the population
\item A way of mixing fragments of good solutions to develop new, potentially even better solutions 
\item A mutation operator to avoid loss of diversity and thus local extrema within the solutions 
\end{itemize}

The advantages of such evolutionary algorithms are, amongst others, the conceptual simplicity because no initial solution or gradient information is required, and their ability to find near-optimal solutions through extensive exploration. Due to their advantages, evolutionary algorithms have been extensively applied to multidisciplinary design optimizations of space transportation systems (e.g., \cite{son2014, marcus2017}). Castellini \cite{castellini2012a} compared genetic algorithms with different state-of-the-art optimization algorithms for multi-objective global optimization. An alternative to genetic algorithms is, for example, a Bayesian-based optimizations \cite{pelamatti2020}.

The Python package DEAP \cite{DEAP_JMLR2012} provides the implementation of the genetic algorithm (called eaSimple). We use the following hyperparameters: population size of \num{5000}, \num{50} generations until termination, a tournament size of the selection
process of 3, and the four probabilities for crossover  and mutation:
a mating probability of \num{0.3}, a mutation probability \num{0.1}, the probability that two corresponding genes are crossed of \num{0.7}, and the probability that one gene mutates of \num{0.5}. A hyperparameter study (see \cite{jentzsch2020}) verified that these hyperparameters lead to robust and converged solutions.

\section{Results of the Optimization Tool}
\label{s:results}

This chapter shows the results of the optimization tool for a typical mission scenario launching a payload into a low-earth orbit (LEO) and into a geostationary transfer orbit (GTO). This study investigates how different propellant combinations and objective functions change the launcher's optimal design. We examine multiple propellant choices for each stage (LOX/LH2, LOX/RP-1, and LOX/CH4) and objective functions (gross lift-off weight (GLOW), structural mass (SM), and a newly defined expendable structural mass (EM)). Furthermore, we want to show how the mission scenario and the optimization goal influence the optimal propellant choice and the rocket's overall design.

\subsection{Missions}
\label{s:mission}
\sisetup{per-mode=symbol}
LEO and GTO orbits are the most requested orbits for current and future missions. Objects in low earth orbit circle earth at an altitude of 200-1000 km. Because of its proximity to the earth's surface, this orbit is commonly used by earth observation satellites and space stations, such as the ISS. In order to stay in orbit, objects need to travel at a speed of around 7.8 km/s. The geostationary transfer orbit is a highly eccentric orbit in which the payload is placed whose target orbit is a geostationary orbit. Geostationary transfer orbits have a perigee (point closest to earth) of 200 km and an apogee (point farthest away from earth) of 35786 km. When the satellite reaches the apogee, it needs to fire its engines to raise the perigee and reach the geostationary orbit. 

After choosing the target orbit, the mission's delta-v budget can be calculated. The delta-v budget or mission velocity is the sum of all flight velocity increments needed to accomplish the mission objective. It is a convenient way to describe the magnitude of the space mission's energy requirement. The ideal total energy requirement to put an object from earth's surface into orbit is the sum of the kinetic orbit energy and the potential energy needed to move the object in the earth's gravitational field from its position on the surface to its orbital altitude. Because of gravity losses (\SIrange{1000}{1500}{\meter \per \second}), aerodynamic drag (\SIrange{100}{150}{\meter \per \second}), maneuvers (\SI{15}{\meter \per \second}) and safety margins (\SIrange{1}{2}{\percent}), the actually required mission velocity is higher than the ideal total delta-v. These losses are difficult to compute since drag, acceleration due to gravity, and flight path angle are unknown functions of time. Experience and data from previous missions provide a basis for conservative values of these losses. 

Furthermore, the rocket gains a velocity increment during launch due to the earth's rotational speed. This effect depends on the latitude of the launch site (\SI{460}{\meter \per \second} for Kourou). Delta-v requirements due to inclination changes are not considered. Table \ref{tab:tabMissions} summarizes the delta-v budget including all losses, and the payload masses for the missions into a GTO of \SI{200}{\kilo \meter} x \SI{35786}{\kilo \metre} and into a LEO of \SI{200}{\kilo \meter} x \SI{200}{\kilo \metre}.

\renewcommand{\arraystretch}{1.6}
\begin{table}[h]
        \centering
        \begin{tabular}{l  c  c}
            \hline\hline
                  \textbf{Parameter} & \textbf{GTO}   & \textbf{LEO}\\
            \hline
           $\Delta v_\text{ideal}$     & \SI{10430}{\metre \per \second}                     & \SI{8030}{\metre \per \second}      \\
          \hline
           $\Delta v_\text{total}$     & \SI{12000}{\metre \per \second}                           & \SI{9500}{\metre \per \second}       \\
          \hline
           $m_\text{pl}$         & \SI{5000}{\kilo \gram}        & \SI{15600}{\kilo \gram} \\
 \hline \hline
	\end{tabular}
	\caption{Mission and Payload Requirements}
	\label{tab:tabMissions}
\end{table}

\subsection{Constraints}
All considered launch vehicles are partially reusable with a first stage that lands via retro-propulsion and an expendable upper stage. All engines use the gas-generator cycle and one of the following propellant combinations: LOX/LH2, LOX/RP-1, LOX/LCH4. Furthermore, we consider mixed-propellant rockets with liquid hydrogen as fuel in the upper stage and one of the hydrocarbons in the first stage. LOX is always used as the oxidizer, so it is no longer explicitly mentioned to distinguish the different fuels from here on. Table \ref{tab:tabAnnexBoundaries} in the appendix shows the parameter ranges and boundary constraints that are applied during the optimization. Apart from these restrictions, the optimization algorithm can change engine parameters such as combustion chamber pressure, mixture ratio, throat diameter, and expansion ratio within predefined limits to find the best possible solution.


\subsection{GLOW}
\begin{figure}[htpb]
  \begin{subfigure}[b]{0.49\textwidth}
    \includegraphics[width=\textwidth]{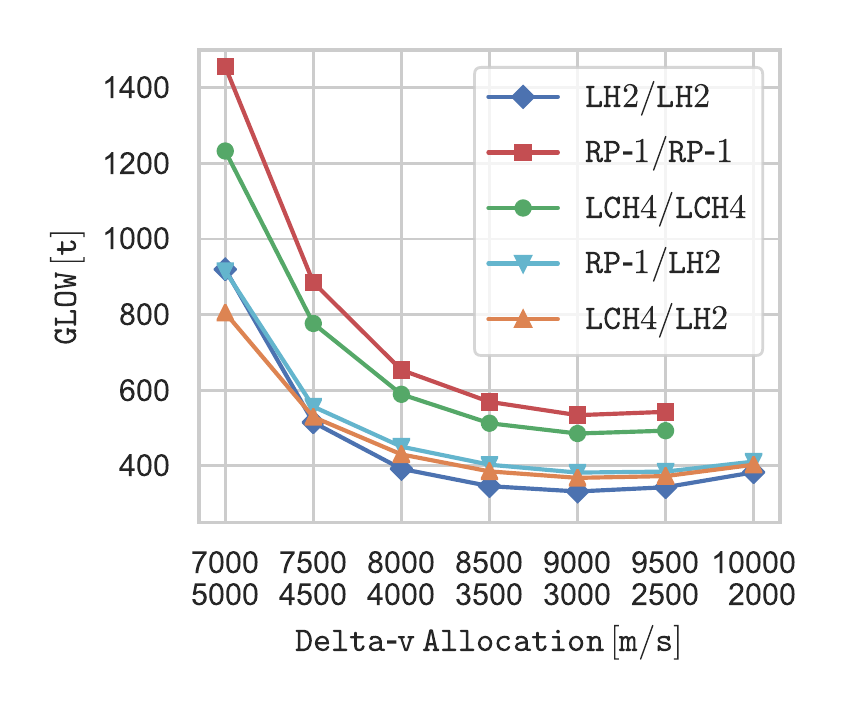}
    \caption{GTO Mission}
    \label{fig:fig3.1}
  \end{subfigure}
  \hfill
  \begin{subfigure}[b]{0.49\textwidth}
    \includegraphics[width=\textwidth]{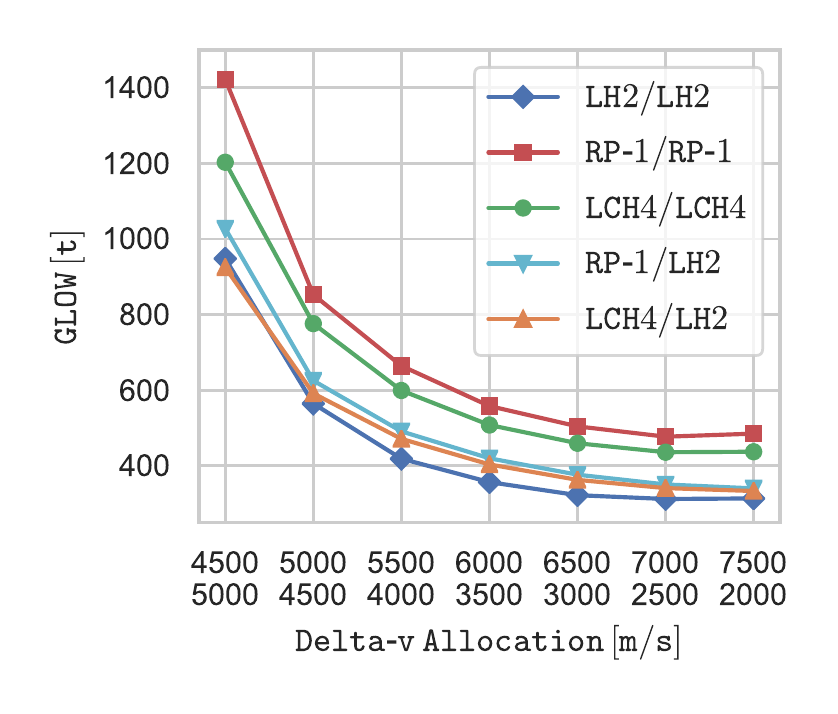}
    \caption{LEO Mission}
    \label{fig:fig3.2}
  \end{subfigure}
  \caption[Comparison of the optimized GLOW]{Comparison of the optimized GLOW. The engines for an extreme delta-v allocation of 2000/10000 \si{\metre \per \second} are beyond the scope of the engine mass estimates; thus, they are discarded.}
  \label{fig:fig3}
\end{figure}

The objective function in this section is the gross lift-off weight (GLOW). Fig. \ref{fig:fig3} shows the minimum GLOW as a function of the delta-v allocation between the first and second stage. For the GTO mission, the minimum GLOW is achieved for a delta-v allocation of around \SI[per-mode=symbol]{3000}{\metre \per \second} for the first stage and \SI[per-mode=symbol]{9000}{\metre \per \second} for the upper stage. For the LEO mission, however, smaller first stages are favorable, and the optimum delta-v allocation differs between different propellant combinations. 

Furthermore, the GLOW explodes for large first stages, which produce more than \SI[per-mode=symbol]{4000}{\metre \per \second}, because more propellant is needed for the reentry burn and the landing maneuvers. Based on these results, we can derive an important conclusion: The optimum delta-v allocation that leads to the lowest GLOW depends on the mission scenarios and can vary between different propellant combinations. As the delta-v allocation plays an essential role during the pre-design, it must be included as a parameter in the optimization routine. It cannot be chosen in advance with a default value.

The minimum GLOW of the launch vehicles shows a similar trend for both missions: The optimized LH2 launch vehicles are significantly lighter than the RP-1 and LCH4 launch vehicles. While showing a similar trend, the LCH4 launch vehicles are around \SI{10}{\percent} lighter than the  RP-1 launch vehicles, with the difference becoming greater for larger first stages. Although they are slightly heavier, the mixed-propellant launch vehicles almost reach the performance of the pure LH2 launcher. This performance is the result of the high specific impulse of the liquid hydrogen upper stage combined with the high-density hydrocarbon fuel in the first stage. Quantitatively speaking, the minimum GLOW for the GTO mission are in increasing order: LH2 \SI{332}{\tonne}, LCH4/LH2 \SI{368}{\tonne}, RP-1/LH2 \SI{382}{\tonne}, LCH4 \SI{485}{\tonne}, and RP-1, \SI{534}{\tonne}.

\subsection{Structural Mass}
\begin{figure}[hbt]
  \begin{subfigure}[b]{0.49\textwidth}
    \includegraphics[width=\textwidth]{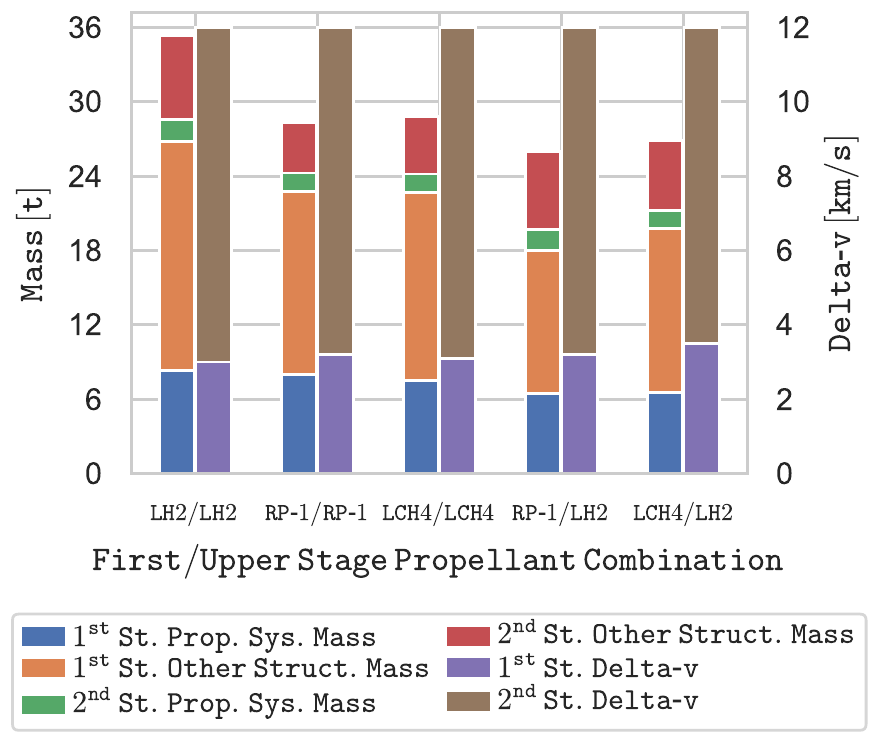}
    \caption{GTO Mission}
    \label{fig:fig3.3}
  \end{subfigure}
  \hfill
  \begin{subfigure}[b]{0.49\textwidth}
    \includegraphics[width=\textwidth]{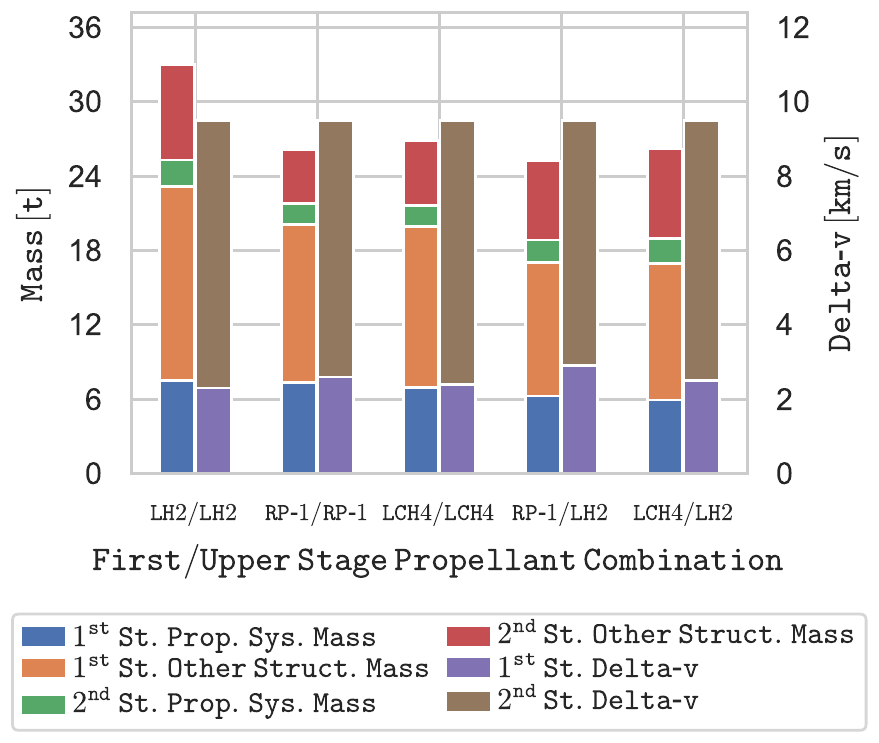}
    \caption{LEO Mission}
    \label{fig:fig3.4}
  \end{subfigure}
  \caption[Structural Mass Breakdown and Delta-v Allocation Comparison]{The figure shows the delta-v allocation and optimized structural mass, which is broken down into the propulsion system mass (engine and thrust-vector control system), and the residual structural mass of each stage.}
  \label{fig:fig3.55}
\end{figure}

The mixed-propellant launch vehicles yield the lowest overall structural mass for both missions, with RP-1/LH2 being the lightest (see Fig.~\ref{fig:fig3.55}). By benefiting from the high $I_{sp}$ of liquid hydrogen and the high density of the hydrocarbons, the mixed-propellant launchers outperform launchers with the same propellant combination in both stages. Regarding the single propellant combinations, LH2 yields the highest structural mass, followed by LCH4 and RP-1. 

A detailed examination of the different launcher designs reveals interesting insights: Comparing both mixed-propellant launchers for the GTO mission, the launch vehicle with RP-1/LH2 has a delta-v allocation of 3200/8800 \si{\meter \per \second} compared to an allocation of 3500/8500 \si{\meter \per \second} for the LCH4/LH2 launcher. These different stage sizes lead to a slightly lower total structural mass of the RP-1/LH2 launcher, while the LCH4/LH2 launch vehicle' larger first stage features a larger portion of the structural mass being reused. For the LEO mission, however, the RP-1/LH2 launcher has both a lower total structural mass and a larger reusable first stage. Fig.~\ref{fig:fig3.55} confirms that the necessary tank insulation causes the much heavier structural masses for cryogenic LH2 stages. Finally, the propulsion system makes up around \SI{20}{\percent} (LH2) to \SI{27}{\percent} (RP-1) of structural mass in the upper stage, and \SI{23}{\percent} (LH2) to \SI{28}{\percent} (RP-1) in the first stage.

By looking at the delta-v allocations, one can see that the upper stage contributes a larger proportion to the total delta-v for the GTO mission. Two possible explanations can be found for this: First, making a reusable first stage even larger and lager increases the total launcher's mass because the required propellant for the reentry burn and landing maneuvers increases drastically. Hence, the upper stage needs to contribute a larger proportion for high delta-v missions such as to a GTO. Secondly, a launcher with a more powerful upper stage benefits from the high ISP of the vacuum optimized upper stage; thus, it can generate more delta-v.

\begin{figure}[htb]
  \begin{subfigure}[b]{0.49\textwidth}
    \includegraphics[width=\textwidth]{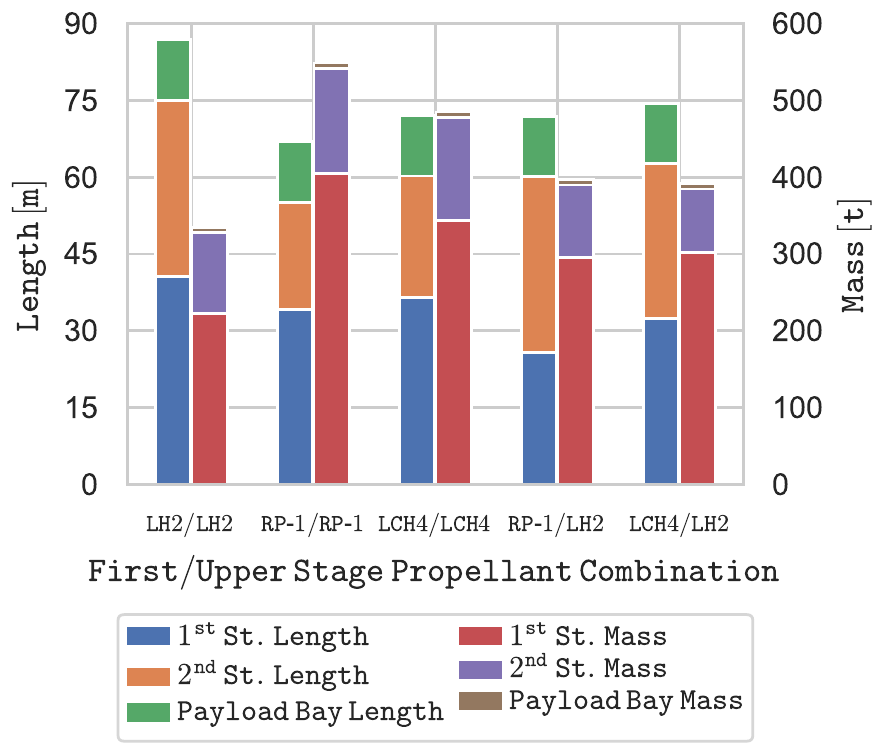}
    \caption{GTO Mission}
    \label{fig:fig3.5}
  \end{subfigure}
  \hfill
  \begin{subfigure}[b]{0.49\textwidth}
    \includegraphics[width=\textwidth]{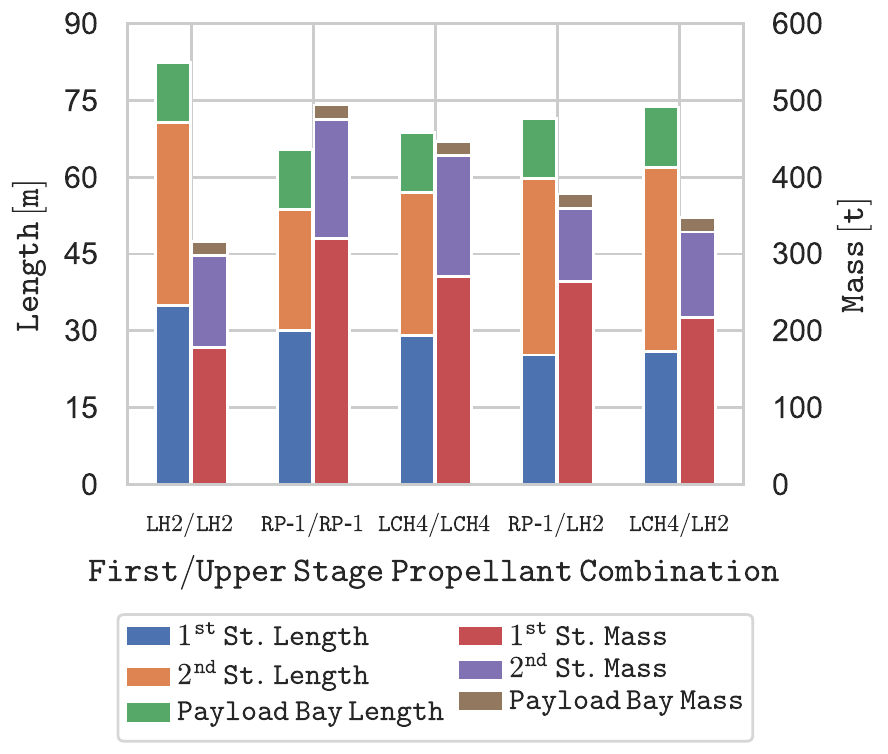}
    \caption{LEO Mission}
    \label{fig:fig3.6}
  \end{subfigure}
  \caption{Length and GLOW Breakdown Comparison}
  \label{fig:fig4}
\end{figure}

Fig.~\ref{fig:fig4} shows the length and GLOW of the optimized launch vehicles for the GTO and the LEO mission broken down into the first stage, upper stage, and payload bay. In general, the LH2 launch vehicles have the lowest GLOW, but they are \SIrange{10}{30}{\percent} longer than the LCH4 launch vehicles, which
in turn are up to \SI{8}{\percent} longer than the RP-1 rockets. This effect is mainly caused by the density differences between the fuels resulting in different sized tanks.

The mixed-propellant RP-1/LH2 and LCH4/LH2 launch vehicles seem to combine the single propellant rocket's advantages, yielding a relatively compact design with moderate GLOW. Comparing the mixed-propellant rockets with the LH2 design for the GTO mission, their length decreases (by up to \SI{15}{\metre}), whereas the GLOW increases only slightly (by up to \SI{60}{\tonne}). In contrast to the RP-1 launch vehicle, the mixed-propellant rocket's length increases by roughly \SI{5}{\metre}, but their GLOW decreases by \SI{150}{\tonne}. In summary, the mixed-propellant rockets present a significant reduction in GLOW compared with the launch vehicles using only RP-1 and LCH4, and a substantial decrease in length compared to the LH2 vehicle. For the mixed-propellant rockets, LCH4/LH2 yields the overall lighter vehicle in terms of GLOW, while RP-1/LH2 has a slight size advantage. The LEO mission shows the same tendencies as the GTO mission, although the differences between individual propellant combinations vary slightly.

\subsection{Expendable Structural Mass}

\begin{figure}[p]
  \begin{subfigure}[b]{0.46\textwidth}
    \includegraphics[width=\textwidth]{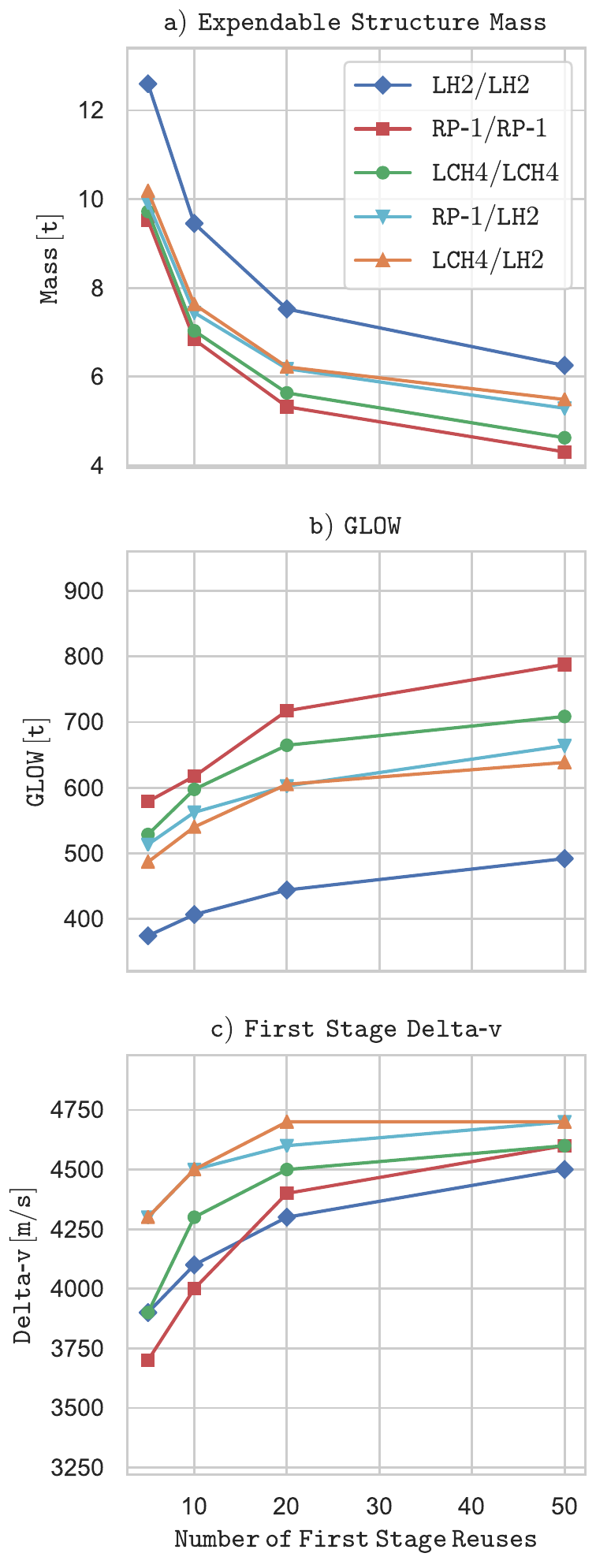}
    \caption{GTO Mission}
    \label{fig:fig3.7}
  \end{subfigure}
  \hfill
  \begin{subfigure}[b]{0.46\textwidth}
    \includegraphics[width=\textwidth]{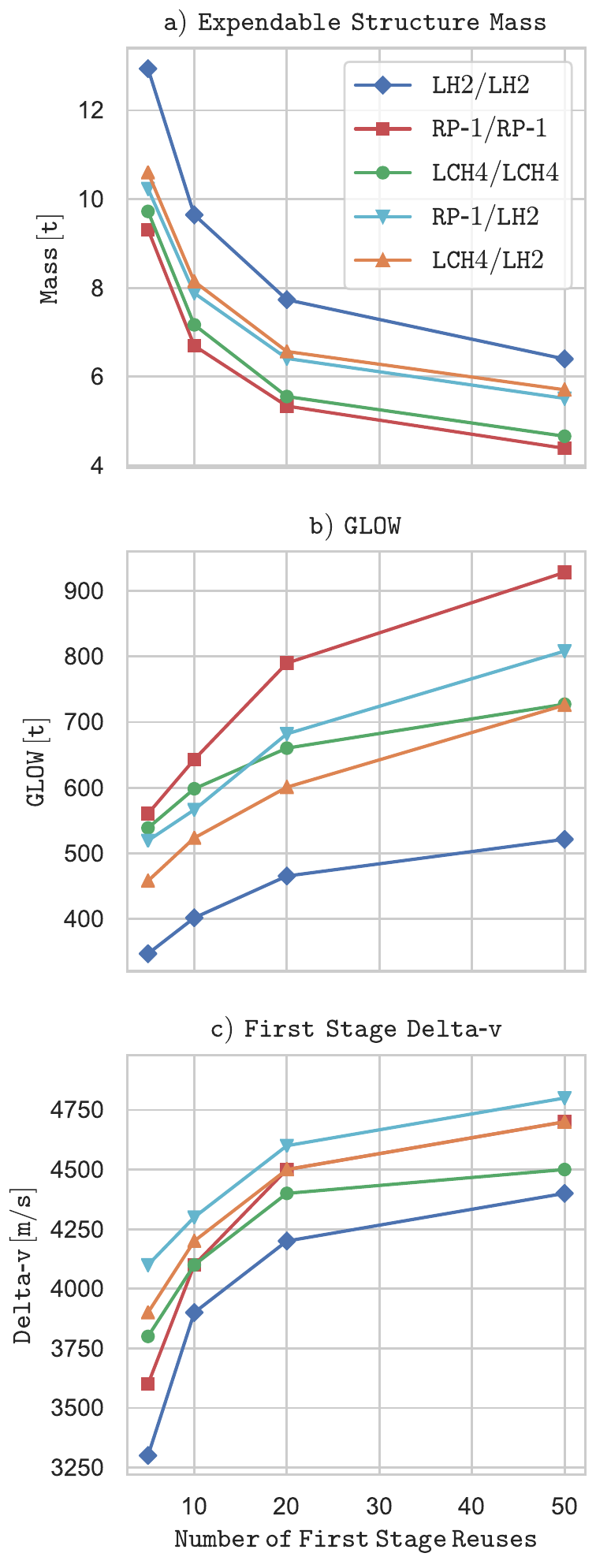}
    \caption{LEO Mission}
    \label{fig:fig3.8}
  \end{subfigure}
  \caption{Expendable Structural Mass, GLOW and First Stage Delta-v Comparison}
      \label{fig:fig3.88}
\end{figure}

Although the previously used optimization objective does achieve the goal of reducing the total structural mass, representing the primary cost driver of rockets, it does not consider reusability. Therefore, we study the expandable structural mass (EM) that consists of the upper stage's structural mass and a fraction of the first stage's structural mass, depending on the number of reuses denoted by $n_\text{reuses}$. For example, if the first stage is reused ten times, \SI{10}{\percent} of the first stage's structural mass is assumed to be expendable. The expendable structure mass can be written as

\begin{equation}
m_\text{EM}= m_\text{s,2} + m_\text{s,1} / n_\text{reuses}.
\end{equation}

Because costs for recovery and refurbishment are not part of this study, the most cost-effective configuration may differ from the optimization program results. However, if these costs can be related to the structural mass, the results can be easily adapted. Four scenarios with 5, 10, 20, and 50 reuses of the first stage are investigated for the same five propellant combination configurations as in the previous sections. For each propellant combination and reuse case, the expendable structural mass, the GLOW, and the first stage delta-v are shown in Fig.~\ref{fig:fig3.88}.

For both mission scenarios and all number of first stage reuses, the LH2 launch vehicles have the highest and the RP-1 launch vehicles the lowest expandable structural mass. The mixed-propellant rockets present values near the hydrocarbon rockets for fewer reuses and are in between the hydrocarbon rockets and the LH2 launch vehicle for a larger number of reuses. This shift occurs because LH2 upper stages have greater structural masses, and the influence of the upper stage on the expendable mass increases with the number of first stage reuses. 

On the other hand, the low expendable structural mass of the hydrocarbon launchers results in very massive designs with a high GLOW, which might be unfavorable. When designing an actual new launch vehicle, a compromise would have to be found, which yields a low expendable structural mass with still moderate GLOW. This is particularly true for the LEO mission with the lower delta-v demands. For the GTO mission, the first stage size is limited; otherwise, the high delta-v mission requirements can no longer be met. 

Increasing the number of reuses from 5 to 10, 20, or 50 yields a decrease in expendable mass of 25-28$\%$, 38-44$\%$, and 47-53$\%$, respectively. Looking at the delta-v allocation of the optimized launch vehicles, a tendency to higher first stage delta-v's with an increasing number of reuses is observable. If the first stage is reused more often, it becomes profitable to build larger first stages and smaller upper stages. However, this entails more massive overall rockets because the larger first stages have a much lower $I_{sp}$ than the vacuum optimized upper stages.

\subsection{Sensitivity Analyses}
By conducting sensitivity analyses, we want to broaden this paper's conclusions to real-world applications, which certainly differ slightly from our modeling framework. Fig. \ref{fig:sensitivity} in the appendix shows the impact of changing delta-v budgets and specific impulses for a \SI{5000}{\kilo \gram} payload GTO mission. As expected, the delta-v budget and the engines' specific impulse significantly affect the GLOW, but the overall trends remain about the same. This shows that this study's general insights remain valid even though components or estimates are modeled in a simplified way.

Furthermore, slight deviations between the graphs - for example, the optimal delta-v distribution for LCH4 shifts depending on the $I_{sp}$ (Fig. \ref{fig:figA.1}) - show that certain parameters must not be defined in advance with traditional values. The optimization algorithm must have the freedom to select these parameters freely, such as the delta-v distribution, to obtain the optimal rocket design.

\section{Discussion}
\label{s:discussion}

So far, we have compared different propellant combinations for two mission scenarios and three different optimization objectives. The results showed that the optimization objective influences the optimum propellant choice and the overall launcher design, weight, and size. This section provides a more in-depth view of chosen design variables for different optimization objectives and propellant combinations. The goal is to show how the launcher's design and engine parameter change for different objective functions. Table \ref{tab:tab4.1} and \ref{tab:tab4.2} compares the optimum launch vehicles for the GTO mission with a \SI{5}{\tonne} payload. For the expendable structural mass, 20 reuses of the first stage are assumed. The payload bay contains the masses of the payload, the fairing, and the avionics bay.

\setlength{\tabcolsep}{3.7pt}
\renewcommand{\arraystretch}{1.58}
\begin{table}[!h]
        \centering
        \begin{tabular}{| c | c | l  c  c  c @{\hskip 12pt}  c  c  c @{\hskip 12pt}  c  c  c|}
            
            \hline\hline
            &   & Prop. Combination & \multicolumn{3}{c@{\hskip 12pt}}{\textbf{LOX/LH2}}  & \multicolumn{3}{c@{\hskip 12pt}}{\textbf{LOX/RP-1}}  & \multicolumn{3}{c|}{\textbf{LOX/LCH4}}  \\[-1ex] 
            &   & Objective Function          & \textbf{GLOW} & \textbf{SM}  & \textbf{EM} & \textbf{GLOW}  & \textbf{SM} & \textbf{EM} & \textbf{GLOW}  & \textbf{SM} & \textbf{EM}   \\
            \hline\hline 
            
            \multirow{7}{*}{\rotatebox[origin=c]{90}{\textbf{Mass and Geometry}}} 
            & \multirow{3}{*}{\rotatebox[origin=c]{90}{\textbf{}}} 
                & {Payload Bay Mass [t]}            & 7.4       & 7.4       & 7.4       & 7.3       & 7.3       & 7.4       & 7.3       & 7.3       & 7.4   \\[1ex]
          
            & \multirow{3.0}{*}{\rotatebox[origin=c]{90}{\textbf{Upper Stage }}}    
                & {Struct. Mass [t]}                & 9.1       & 8.6       & 5.6       & 5.9       & 5.5       & 3.8       & 7.0       & 6.1       & 4.1   \\ [-1.5ex]
            &   & {Prop. Mass [t]}                  & 101.1     & 95.9      & 54.6      & 143.4     & 131.3     & 74.9      & 154.0     & 127.4     & 67.9  \\ [-1.5ex]
            &   & {Struct. Coeff. [-]}              & 0.082     & 0.082     & 0.093     & 0.039     & 0.040     & 0.049     & 0.043     & 0.046     & 0.057 \\ [-1.5ex]
            &   & {Diameter [m]}                    & 3.8       & 3.6       & 3.8       & 4.0       & 3.8       & 3.4       & 4.2       & 3.8       & 3.6   \\ [1ex]
           
            & \multirow{4.0}{*}{\rotatebox[origin=c]{90}{\textbf{First Stage}}}
                & {Struct. Mass [t]}                & 27.4      & 26.8      & 38.3      & 23.0      & 22.8      & 30.0      & 22.3      & 22.7      & 30.9  \\ [-1.5ex]
            &   & {Tot. Prop. Mass [t]}             & 182.8     & 196.2     & 337.9     & 350.9     & 381.6     & 601.1     & 288.6     & 321.3     & 554.4 \\ [-1.5ex]
            &   & {Land. Prop. Mass [t]}            & 11.7      & 13.0      & 40.5      & 15.0      & 17.8      & 49.0      & 10.5      & 15.3      & 49.1  \\ [-1.5ex]
            &   & {Struct. Coeff. [-]}              & 0.130     & 0.120     & 0.102     & 0.062     & 0.056     & 0.048     & 0.072     & 0.066     & 0.053 \\ [-1.5ex]
            &   & {Diameter [m]}                    & 4.4       & 4.4       & 5.0       & 4.2       & 4.2       & 4.4       & 4.2       & 4.2       & 4.6   \\
           
            \hline
            \multirow{10.5}{*}{\rotatebox[origin=c]{90}{\textbf{Propulsion System}}} 
            & \multirow{5}{*}{\rotatebox[origin=c]{90}{\textbf{Upper Stage}}}
                & {Number of Engines}               & 1         & 1         & 1         & 1         & 1         & 1         & 1         & 1         & 1     \\ [-1.5ex]
            &   & {F$_\text{vac}$ [kN]}                  & 1095      & 1028      & 614       & 1439      & 1341      & 790       & 1574      & 1298      & 742   \\ [-1.5ex]
            &   & {I$_{sp\text{\_vac}}$ [s]}               & 450       & 448       & 445       & 353       & 352       & 354       & 367       & 366       & 368   \\ [-1.5ex]
            &   & {t$_\text{b}$ [s]}                     & 408       & 410       & 389       & 345       & 338       & 329       & 352       & 353       & 330   \\ [-1.5ex]
            &   & {p$_\text{c}$ [bar]}                   & 115       & 100       & 85        & 110       & 110       & 90        & 105       & 100       & 85    \\ [-1.5ex]
            &   & {ROF$_{\text{engine}}$ [-]}                   & 6.5       & 6.7       & 7.0       & 2.3       & 2.4       & 2.5       & 3.1       & 3.1       & 3.2   \\ [-1.5ex]
            &   & {$\varepsilon$ [-]}               & 200       & 200       & 200       & 200       & 190       & 195       & 195       & 180       & 200   \\ [1ex]
    
            & \multirow{5.7}{*}{\rotatebox[origin=c]{90}{\textbf{First Stage}}}
                & {Number of Engines}               & 5         & 5         & 5         & 6         & 5         & 6         & 5         & 5         & 5     \\ [-1.5ex]
            &   & {F$_\text{vac, tot}$ [kN]}                  & 4896      & 4905      & 6407      & 7717      & 7773      & 10488     & 7032      & 6958      & 9956  \\ [-1.5ex]
            &   & {I$_{sp\text{\_vac}}$ [s]}               & 418       & 399       & 409       & 319       & 307       & 315       & 334       & 327       & 335   \\ [-1.5ex]
            &   & {I$_{sp\text{\_sl}}$ [s]}                & 369       & 362       & 361       & 281       & 286       & 283       & 292       & 298       & 290   \\ [-1.5ex]
            &   & {t$_\text{b}$ [s]}                     & 153       & 157       & 212       & 143       & 148       & 177       & 134       & 148       & 183   \\ [-1.5ex]
            &   & {p$_\text{c}$ [bar]}                   & 115       & 115       & 135       & 110       & 120       & 130       & 105       & 120       & 135   \\ [-1.5ex]
            &   & {ROF$_{\text{engine}}$ [-]}                   & 5.5       & 6.7       & 6.4       & 2.1       & 2.0       & 2.0       & 2.8       & 2.8       & 2.8   \\ [-1.5ex]
            &   & {$\varepsilon$ [-]}               & 25        & 20        & 30        & 25        & 15        & 25        & 25        & 20        & 35    \\
               
            \hline
            \multirow{3.5}{*}{\rotatebox[origin=c]{90}{\textbf{Total}}}
            &   & {Delta-v Alloc. [km/s]}       & 2.9/9.1   & 3.0/9.0   & 4.3/7.7   & 3.0/9.0   & 3.2/8.8   & 4.4/7.6   & 2.7/9.3   & 3.1/8.9   & 4.5/7.5   \\ [-1.5ex]
            &   & {Length [m]}                      & 87.8      & 86.8      & 87.3      & 65.5      & 66.9      & 75.4      & 70.3      & 72.0      & 79.8  \\ [-1.5ex]
            &   & {Tot. Struct. Mass [t]}           & 36.4      & 35.4      & 43.9      & 28.9      & 28.3      & 33.8      & 29.3      & 28.8      & 35.0  \\ [-1.5ex]
            &   & {Expendable Mass [t]}             & 10.5      & 9.9       & 7.5       & 7.1       & 6.6       & 5.3       & 8.1       & 7.2       & 5.6   \\ [-1.5ex]
            &   & {GLOW [t]}                        & 327.8     & 334.9     & 443.9     & 530.6     & 548.5     & 717.2     & 479.2     & 484.9     & 664.7 \\
            \hline \hline
            
	\end{tabular}
	\caption{Optimized Launch Vehicles with the Same Propellant Combination in Both Stages}
	\label{tab:tab4.1}
\end{table}

\setlength{\tabcolsep}{3.7pt}
\renewcommand{\arraystretch}{1.6}
\begin{table}[!h]
        \centering
        \begin{tabular}{| c | c | l  c  c  c @{\hskip 12pt}  c  c  c |}
            
            \hline\hline
            &   & Prop. Combination & \multicolumn{3}{c@{\hskip 12pt}}{\hskip -5pt \textbf{LOX/RP-1 $\&$ LOX/LH2}}  & \multicolumn{3}{c|}{\hskip -8pt \textbf{LOX/LCH4 $\&$ LOX/LH2}} \\[-1ex] 
            &   & Objective Function            & \textbf{GLOW} & \textbf{SM}  & \textbf{EM} & \textbf{GLOW}  & \textbf{SM} & \textbf{EM} \\
            \hline\hline 
            
            \multirow{8}{*}{\rotatebox[origin=c]{90}{\textbf{Mass and Geometry}}} 
            & \multirow{3}{*}{\rotatebox[origin=c]{90}{\textbf{}}} 
                & {Payload Bay Mass [t]}            & 7.3       & 7.4       & 7.4       & 7.3       & 7.4       & 7.4   \\[1ex]
          
            & \multirow{3.0}{*}{\rotatebox[origin=c]{90}{\textbf{Upper Stage }}}    
                & {Struct. Mass [t]}                & 9.8       & 7.9       & 4.9       & 9.0       & 7.2       & 4.8   \\ [-1.5ex]
            &   & {Prop. Mass [t]}                  & 110.8     & 86.7      & 46.6      & 100.0     & 75.4      & 44.9  \\ [-1.5ex]
            &   & {Struct. Coeff. [-]}              & 0.081     & 0.084     & 0.095     & 0.083     & 0.087     & 0.096 \\ [-1.5ex]
            &   & {Diameter [m]}                    & 3.8       & 3.4       & 3.4       & 3.8       & 3.4       & 3.4   \\ [1ex]
           
            & \multirow{4.0}{*}{\rotatebox[origin=c]{90}{\textbf{First Stage}}}
                & {Struct. Mass [t]}                & 17.6      & 18.0      & 25.7      & 18.8      & 19.7      & 28.8  \\ [-1.5ex]
            &   & {Tot. Prop. Mass [t]}             & 235.4     & 277.0     & 517.9     & 232.9     & 282.3     & 519.2 \\ [-1.5ex]
            &   & {Land. Prop. Mass [t]}            & 8.7       & 14.0      & 46.2      & 10.6      & 17.8      & 50.5  \\ [-1.5ex]
            &   & {Struct. Coeff. [-]}              & 0.070     & 0.061     & 0.047     & 0.075     & 0.065     & 0.053 \\ [-1.5ex]
            &   & {Diameter [m]}                    & 4.0       & 4.2       & 4.4       & 4.4       & 4.2       & 4.6   \\
           
            \hline
            \multirow{10.5}{*}{\rotatebox[origin=c]{90}{\textbf{Propulsion System}}} 
            & \multirow{5}{*}{\rotatebox[origin=c]{90}{\textbf{Upper Stage}}}
                & {Number of Engines}               & 1         & 1         & 1         & 1         & 1         & 1     \\ [-1.5ex]
            &   & {F$_\text{vac}$ [kN]}                  & 1186      & 946       & 538       & 1075      & 841       & 522   \\ [-1.5ex]
            &   & {I$_{sp\text{\_vac}}$ [s]}               & 450       & 448       & 445       & 450       & 448       & 446   \\ [-1.5ex]
            &   & {t$_\text{b}$ [s]}                     & 413       & 403       & 378       & 411       & 394       & 376   \\ [-1.5ex]
            &   & {p$_\text{c}$ [bar]}                   & 115       & 100       & 75        & 105       & 115       & 80    \\ [-1.5ex]
            &   & {ROF$_{\text{engine}}$ [-]}                   & 6.4       & 6.7       & 7.1       & 6.4       & 6.7       & 7.0   \\ [-1.5ex]
            &   & {$\varepsilon$ [-]}               & 200       & 195       & 200       & 200       & 195       & 200   \\ [1ex]
    
            & \multirow{5.7}{*}{\rotatebox[origin=c]{90}{\textbf{First Stage}}}
                & {Number of Engines}               & 5         & 5         & 5         & 5         & 5         & 5     \\ [-1.5ex]
            &   & {F$_\text{vac, tot}$ [kN]}                  & 5691      & 5504      & 8654      & 5454      & 5547      & 8861  \\ [-1.5ex]
            &   & {I$_{sp\text{\_vac}}$ [s]}               & 321       & 309       & 318       & 337       & 327       & 333   \\ [-1.5ex]
            &   & {I$_{sp\text{\_sl}}$ [s]}                & 279       & 285       & 283       & 291       & 298       & 295   \\ [-1.5ex]
            &   & {t$_\text{b}$ [s]}                     & 130       & 153       & 187       & 141       & 163       & 191   \\ [-1.5ex]
            &   & {p$_\text{c}$ [bar]}                   & 120       & 105       & 120       & 115       & 120       & 115   \\ [-1.5ex]
            &   & {ROF$_{\text{engine}}$ [-]}                   & 2.1       & 2.1       & 2.1       & 2.8       & 2.8       & 2.9   \\ [-1.5ex]
            &   & {$\varepsilon$ [-]}               & 30        & 15        & 25        & 30        & 20        & 25    \\
               
            \hline
            \multirow{3.5}{*}{\rotatebox[origin=c]{90}{\textbf{Total}}}
            &   & {Delta-v Alloc. [km/s]}           & 2.7/9.3   & 3.2/8.8   & 4.6/7.4   & 2.9/9.1   & 3.5/8.5   & 4.7/7.3   \\ [-1.5ex]
            &   & {Length [m]}                      & 73.2      & 71.8      & 74.4      & 71.6      & 74.4      & 79.8  \\ [-1.5ex]
            &   & {Tot. Struct. Mass [t]}           & 27.4      & 26.0      & 30.6      & 27.8      & 26.9      & 33.6  \\ [-1.5ex]
            &   & {Expendable Mass [t]}             & 10.7      & 8.8       & 6.2       & 9.9       & 8.2       & 6.2   \\ [-1.5ex]
            &   & {GLOW [t]}                        & 381.0     & 397.1     & 602.5     & 368.1     & 392.0     & 605.1 \\
            \hline \hline
            
	\end{tabular}
	\caption{Optimized Launch Vehicles with Different Propellant Combinations in Each Stage}
	\label{tab:tab4.2}
\end{table}

Comparing the results for the different objective functions, we can derive implications, which are similar for all propellant combinations: As already observed before, the optimum staging changes for different objective functions. If the rockets are optimized for a minimum expendable mass, the delta-v contribution of the first stage increases, leading to a more massive stage with a larger height or diameter. Consequently, the required propellant for the reentry burn and landing maneuvers grows. With this higher total stage mass, the engines' thrust must also be increased, which simultaneously leads to higher combustion chamber pressures in the first stage engines. 

On the other hand, with larger first stages, the size of the upper stages shrinks as their delta-v demands are much lower. The smaller size leads to lower structural masses, which is favorable as this mass is not recoverable. Similarly, the upper stage engine's thrust and chamber pressure reduce with decreasing demands.

Looking at the structural mass optimized launch vehicles, one can see that the optimization was successful and, in fact, leads to the lowest total structural masses. For LH2, the mixture ratio is increased to benefit from the much higher density of liquid oxygen compared with liquid hydrogen, thus featuring much smaller overall tank volumes and tank masses. For the hydrocarbon propellants, the optimizer chooses engines with a high sea level $I_{sp}$, achieved due to higher combustion chamber pressures or smaller expansion ratios.

Overall, the optimal number of engines in the first stage seems to be relatively small. For the implemented empirically-based mass estimates, a larger number of engines apparently leads to overall higher masses or poorer engine performance, which is unfavorable. For further statements about the optimal number of engines, a more complex mass estimation, based on the masses of the individual engine components, may have to be modeled, since the currently used correlations only consider the engine mass as a function of the engine thrust.

At an engine level, the upper stage's mixture ratio is higher than that of the first stage. In general, the $I_{sp}$ of a gas-generator engine deteriorates slightly for higher mixture ratios, but the propellant tanks can be made much more compact as more high-density LOX can be stored. This effect is notably pronounced for liquid hydrogen or for a structural mass optimized rocket, since the tank mass is particularly relevant in this case. Of course, real technical implementation could have reasons against such high mixture ratios (e.g., cooling, combustion efficiency). Therefore, it could be that one wants to take this into account and prefer a lower mixture ratio. Nevertheless, the trend towards higher mixture ratios seems to be beneficial to the overall launcher performance.

Finally, we want to comment on the structural coefficients: As is generally known, the size of a stage strongly influences the structural index, which tends to decrease when a stage, and thus the tanks and structural components, are built larger. Based on our study, it is also clear that the structural index must not be defined in advance based on traditional values. This would prefer particular configurations and designs because different structural coefficients emerge from the optimization, depending on the optimization goal.

\section{Conclusion and Outlook}
\label{s:conclusion}

In this work, we presented an optimization framework for partially reusable launch vehicles. Those launch vehicles consist of two stages with a reusable first stage landing via retro-propulsion similar to SpaceX's Falcon 9. Using suitable mass estimates of all essential subsystems and a routine to calculate the needed propellant for the ascent and descent, we employed a genetic algorithm to find optimal designs for different objective functions and propellant combinations: LOX/LH2, LOX/RP-1, LOX/LCH4. Concerning propulsion, we assume gas-generator-cycle liquid rocket engines in both stages. Apart from this, the optimization algorithm can change engine parameters such as combustion chamber pressure, mixture ratio, throat diameter, and expansion ratio within predefined limits to find the best possible solution.

Not surprisingly, our results show that the question of which design and which propellant combination is the most suitable depends strongly on the assumed objective function. If one optimizes according to GLOW, the launch vehicles that use LOX/LH2 in both stages perform best. If one optimizes according to structural mass, launch vehicles with LOX/RP-1 or LOX/LCH4 in the first stage and LOX/LH2 in the second stage are optimal. On the other hand, if one optimizes according to an expandable structural mass for five reuses and more, the configurations that only use LOX/RP-1 or LOX/LCH4 give the best results. Furthermore, the optimal engine design parameters also change depending on the chosen objective function.

The present work can be improved in many directions: First, different propellant tank designs like using a common bulkhead should be included, and the tank design should be further optimized (see \cite{vietze2018}). Second, the simplified modeling of aerodynamics and losses should be upgraded. Third, other engine power cycles, e.g., staged-combustion or expander, could be investigated. It would be exciting to study if and how the coupling with a trajectory optimization program changes the results on a global level. 

Finally, to investigate the most relevant goal of cost reduction, appropriate cost estimates must be integrated. Using appropriate cost estimates instead of mass-based objective functions would result in the launcher design that yields the lowest overall cost for a given payload and mission.

\newpage
\section*{Appendix}

\subsection{Comparison with Falcon 9}

\renewcommand{\arraystretch}{1.6}
\setlength{\tabcolsep}{4pt}
\begin{table}[h!]
        \centering
        \begin{tabular}{|c| c| l@{\hskip 3pt}  S@{\hskip 3pt}  S |}
        \hline\hline
     
        &   &                                   & \textbf{Falcon 9}  & \textbf{Optimizer} \\
                
        \hline\hline 
        \multirow{9}{*}{\rotatebox[origin=c]{90}{\textbf{Mass and Geometry}}} &  \multirow{3}{*}{\rotatebox[origin=c]{90}{\textbf{}}} 
        & Payload Bay Mass [t]                      & 7.4         & 7.4    \\[-1.5ex]
        &   & Fairing Length [m]                & 13.2      & 10.7    \\ [1ex]
          
     &  \multirow{3.5}{*}{\rotatebox[origin=c]{90}{\textbf{Upper Stage}}}    
        & Struct. Mass [t]                    & 4.5       & 4.9      \\[-1.5ex]
        &   & Prop. Mass [t]               & 111.5     & 113.7    \\[-1.5ex]
        &   & Struct. Coeff. [-]        & 0.039     & 0.041     \\[-1.5ex]
        &   & Length [m]                        & 16.0      & 20.5      \\[-1.5ex]
        &   & Diameter [m]                      & 3.66      & 3.66      \\[1ex]

     &  \multirow{4.5}{*}{\rotatebox[origin=c]{90}{\textbf{First Stage}}}     
        & Struct. Mass [t]                    & 27.2      & 27.4     \\[-1.5ex]
        &   & Tot. Prop. Mass [t]         & 418.7     & 436.6    \\[-1.5ex]
        &   & Land. Prop. Mass [t]       & 25.0      & 26.6     \\[-1.5ex]
        &   & Struct. Coeff. [-]        & 0.061     & 0.059     \\[-1.5ex]
        &   & Length [m]                        & 40.9      & 48.3      \\[-1.5ex]
        &   & Diameter [m]                      & 3.66      & 3.66      \\
           
        \hline
        \multirow{7.5}{*}{\rotatebox[origin=c]{90}{\textbf{Propulsion System}}} 
     &  \multirow{3.3}{*}{\rotatebox[origin=c]{90}{\textbf{Upper Stage}}}     
        & Number of Engines                     & 1         & 1         \\[-1.5ex]
        &   & F$_\text{vac}$ [kN]                    & 981       & 1074      \\[-1.5ex]
        &   & I$_{sp\text{\_vac}}$ [s]                 & 348       & 351       \\[-1.5ex]
        &   & t$_\text{b}$ [s]                       & 397       & 364       \\ [1ex]

     &  \multirow{4.8}{*}{\rotatebox[origin=c]{90}{\textbf{First Stage}}}     
        & Number of Engines                     & 9         & 9         \\[-1.5ex]
        &   & F$_\text{vac, tot}$ [kN]                    & 8227      & 8536      \\[-1.5ex]
        &   & F$_\text{sl, tot}$ [kN]                     & 7607      & 7770      \\[-1.5ex]
        &   & I$_{sp\text{\_vac}}$ [s]                 & 312       & 310       \\[-1.5ex]
        &   & I$_{sp\text{\_sl}}$ [s]                  & 283       & 282       \\[-1.5ex]
        &   & t$_\text{b}$ [s]                       & 162       & 156       \\
               
        \hline
        \multirow{1.7}{*}{\rotatebox[origin=c]{90}{\textbf{Total}}} 
        &   & GLOW [t]                          & 569.3     & 589.9    \\[-1.5ex]
        &   & Length [m]                        & 70.1      & 80.6      \\
        \hline
        \multirow{2.5}{*}{\rotatebox[origin=c]{90}{\textbf{ Delta-v}}} 
        &   & Upper Stage [km/s]         & 8.5      & 8.5      \\[-1.5ex]
        &   & First Stage [km/s]         & 3.5      & 3.5      \\  [-1.5ex]
        &   & Total [km/s]               & 12.0     & 12.0     \\
 \hline \hline
	\end{tabular}
	\caption{Comparison of Falcon 9 with the results of the reusable launch vehicle framework}
	\label{tab:tabA.1}
\end{table}

\newpage
\subsection{Rocket Parameters - Boundary Values and Constraints}
\renewcommand{\arraystretch}{1.6}
\begin{table}[h]
        \centering
        \begin{tabular}{l  c  c}
            \hline\hline
                \textbf{Parameter}      & \textbf{Boundary Values}   & \textbf{Boundary Values}\\
                                        & \textbf{First Stage}       & \textbf{Upper Stage}\\
            \hline
                Stage Radius        & \SIrange{1.5}{4}{\metre}   & \SIrange{1.5}{4}{\metre}          \\[-1ex]
                                        &                            & min. 0.75 $ r_{1^{st}\text{Stage}}$ \\[-1ex]
                                        &                            & max. 1.00 $
                                        r_{1^{st}\text{Stage}}$  \\
          \hline
           Nozzle Throat Diameter       & \SIrange{0.1}{1}{\metre}  & \SIrange{0.1}{1}{\metre}  \\
          \hline
           Combustion Chamber Pressure  & \SIrange{50}{200}{\bar}   & \SIrange{20}{200}{\bar} \\
          \hline
           Nozzle Area Expansion Ratio  & \numrange{10}{90}         & \numrange{80}{200} \\
          \hline
           Mixture Ratio                & \multicolumn{2}{c}{LOX/LH2: \numrange{4.0}{7.9}}  \\[-1ex]
                                        & \multicolumn{2}{c}{LOX/RP-1: \numrange{1.5}{3.5}}  \\[-1ex]
                                        & \multicolumn{2}{c}{LOX/LCH4: \numrange{2.0}{4.0}}  \\
            \hline
           Minimum Number of Engines     & 5                      & 1 \\
          \hline
           Maximum Number of Engines     & 15                      & 1 \\
          \hline
           Minimum Acceleration         & 1.3 g       & 0.95 g \\
          \hline
           Maximum Length/Diameter    & \multicolumn{2}{c}{20}  \\
 \hline \hline
	\end{tabular}
	\caption{Boundary values and constraints for the optimization program}
	\label{tab:tabAnnexBoundaries}
\end{table}

\newpage
\subsection{Sensitivity Analyses}

\begin{figure}[!h]
  \begin{subfigure}[b]{0.48\textwidth}
    \includegraphics[scale=0.965]{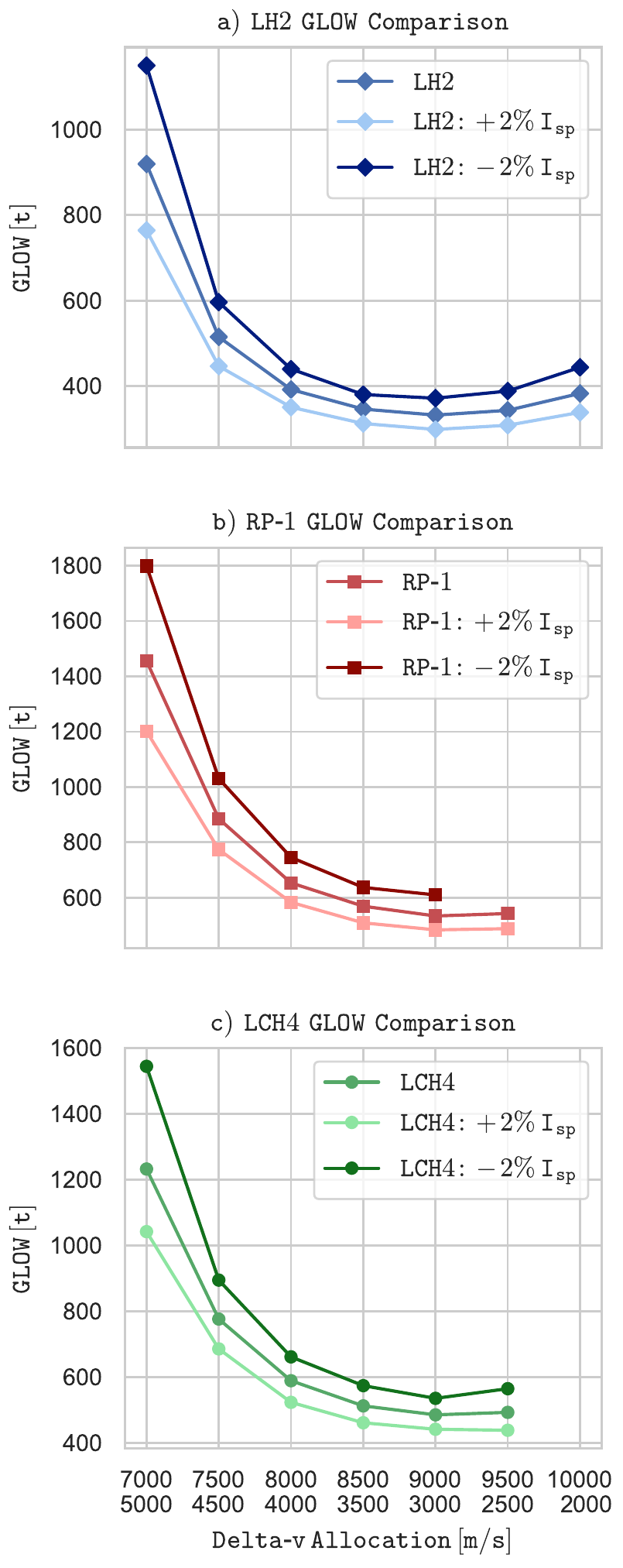}
    \caption{Specific Impulse}
    \label{fig:figA.1}
  \end{subfigure}
  \hfill
  \begin{subfigure}[b]{0.48\textwidth}
    \includegraphics[scale=0.965]{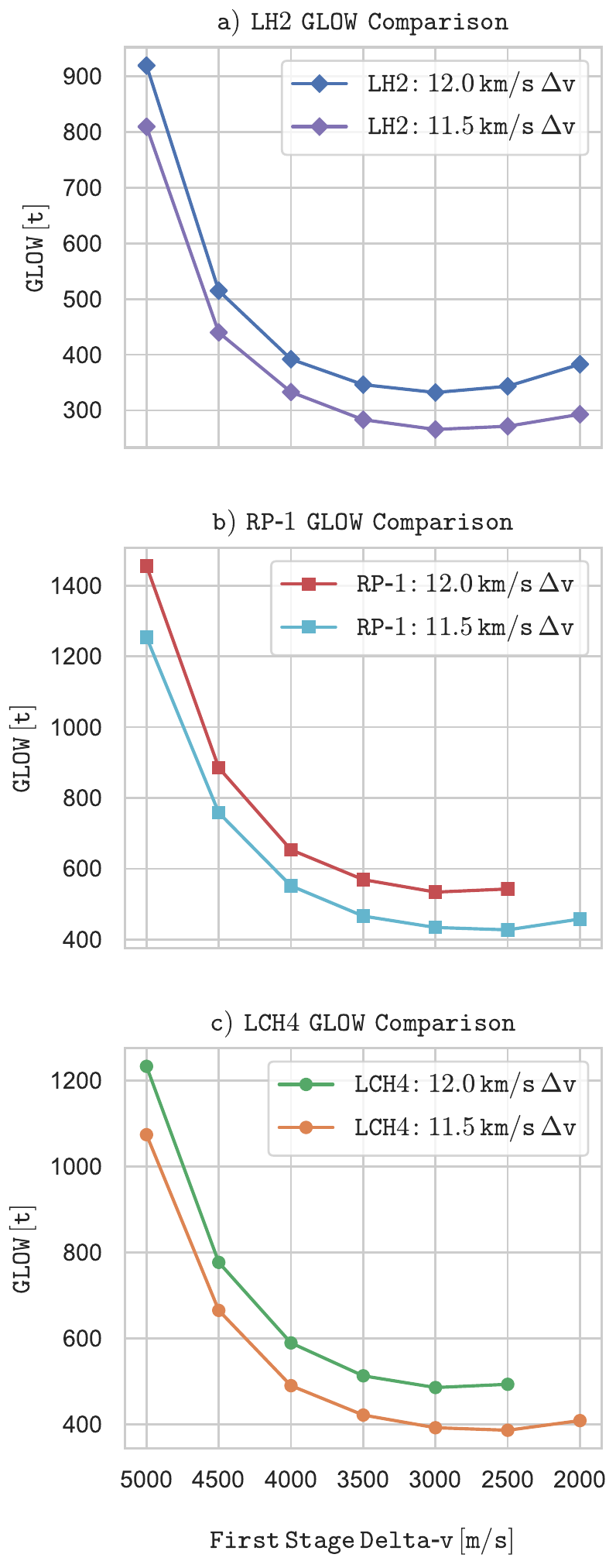}
    \caption{Delta-v Budget}
    \label{fig:figA.2}
  \end{subfigure}
  \caption{Sensitivity Analyses}
   \label{fig:sensitivity}
\end{figure}

\newpage
\bibliography{literature}

\begin{thebibliography}{38}
\newcommand{\enquote}[1]{``#1''}
\providecommand{\natexlab}[1]{#1}
\providecommand{\url}[1]{\texttt{#1}}
\providecommand{\urlprefix}{URL }
\expandafter\ifx\csname urlstyle\endcsname\relax
  \providecommand{\doi}[1]{\discretionary{}{}{}https://doi.org/#1}\else
  \providecommand{\doi}[1]{\discretionary{}{}{}\urlstyle{rm}\url{https://doi.org/#1}}\fi

\bibitem[{Blair et~al.(2001)Blair, Ryan, Schutzenhofer, Signal, and
  Humphries}]{blair2001}
Blair, J.~C., Ryan, R.~S., Schutzenhofer, L.~A., Signal, A., and Humphries,
  W.~R., \enquote{Launch {{Vehicle Design Process}}: {{Characterization}},
  {{Technical Integration}}, and {{Lessons Learned}},} {{NASA Technical
  Publication}}, 2001.

\bibitem[{Braun et~al.(1996)Braun, Moore, and Kroo}]{braun1996}
Braun, R., Moore, A., and Kroo, I., \enquote{Use of the Collaborative
  Optimization Architecture for Launch Vehicle Design,} \emph{6th {{Symposium}}
  on {{Multidisciplinary Analysis}} and {{Optimization}}}, Multidisciplinary
  {{Analysis Optimization Conferences}}, {American Institute of Aeronautics and
  Astronautics}, 1996.
\newblock \doi{10.2514/6.1996-4018}.

\bibitem[{Rowell et~al.(1996)Rowell, Olds, and Unal}]{rowell1996}
Rowell, L., Olds, J., and Unal, R., \enquote{Recent Experiences in
  Multidisciplinary Conceptual Design Optimization for Launch Vehicles,}
  \emph{6th {{Symposium}} on {{Multidisciplinary Analysis}} and
  {{Optimization}}}, Multidisciplinary {{Analysis Optimization Conferences}},
  {American Institute of Aeronautics and Astronautics}, 1996.
\newblock \doi{10.2514/6.1996-4050}.

\bibitem[{Koelle(2000)}]{koelle2000}
Koelle, D., \enquote{The Cost-Optimal Size of Future Reusable Launch Vehicles,}
  \emph{Acta Astronautica}, Vol.~47, No. 2-9, 2000, pp. 205--213.
\newblock \doi{10.1016/S0094-5765(00)00060-6}.

\bibitem[{Durant{\'e} et~al.(2004)Durant{\'e}, Dufour, Pain, Baudrillard, and
  Schoenauer}]{durante2004}
Durant{\'e}, N., Dufour, A., Pain, V., Baudrillard, G., and Schoenauer, M.,
  \enquote{Multi-Disciplinary {{Analysis}} and {{Optimisation Approach}} for
  the {{Design}} of {{Expendable Launchers}},} \emph{10th {{AIAA}}/{{ISSMO
  Multidisciplinary Analysis}} and {{Optimization Conference}}}, {American
  Institute of Aeronautics and Astronautics}, {Albany, New York}, 2004.
\newblock \doi{10.2514/6.2004-4441}.

\bibitem[{Bayley et~al.(2007)Bayley, Hartfield, Burkhalter, and
  Jenkins}]{bayley2007}
Bayley, D., Hartfield, R., Burkhalter, J., and Jenkins, R., \enquote{Design
  {{Optimization}} of a {{Space Launch Vehicle Using}} a {{Genetic
  Algorithm}},} \emph{48th {{AIAA}}/{{ASME}}/{{ASCE}}/{{AHS}}/{{ASC
  Structures}}, {{Structural Dynamics}}, and {{Materials Conference}}},
  {American Institute of Aeronautics and Astronautics}, {Honolulu, Hawaii},
  2007.
\newblock \doi{10.2514/6.2007-1863}.

\bibitem[{Briggs et~al.(2007)Briggs, Ray, and Milthorpe}]{briggs2007}
Briggs, G., Ray, T., and Milthorpe, J., \enquote{Evolutionary {{Algorithm Use}}
  in {{Optimisation}} of a {{Launch Vehicle Stack Model}},} \emph{45th {{AIAA
  Aerospace Sciences Meeting}} and {{Exhibit}}}, {American Institute of
  Aeronautics and Astronautics}, {Reno, Nevada}, 2007.
\newblock \doi{10.2514/6.2007-364}.

\bibitem[{Castellini et~al.(2010)Castellini, Lavagna, Riccardi, and
  Bueskens}]{castellini2010}
Castellini, F., Lavagna, M., Riccardi, A., and Bueskens, C.,
  \enquote{Multidisciplinary {{Design Optimization Models}} and {{Algorithms}}
  for {{Space Launch Vehicles}},} \emph{13th {{AIAA}}/{{ISSMO Multidisciplinary
  Analysis Optimization Conference}}}, {American Institute of Aeronautics and
  Astronautics}, {Fort Worth, Texas}, 2010.
\newblock \doi{10.2514/6.2010-9086}.

\bibitem[{Castellini(2012)}]{castellini2012}
Castellini, F., \enquote{Multidisciplinary {{Design Optimization}} for
  {{Expendable Launch Vehicles}},} Ph.D. thesis, Polytechnic University of
  Milan, 2012.

\bibitem[{Patureau De~Mirand et~al.(2019)Patureau De~Mirand, Bahu, and
  Louaas}]{patureaudemirand2019}
Patureau De~Mirand, A., Bahu, J.-M., and Louaas, E., \enquote{Ariane {{Next}},
  a Vision for a Reusable Cost Efficient {{European}} Rocket,}
  \emph{Proceedings of the 8th {{European Conference}} for {{Aeronautics}} and
  {{Space Sciences}}}, {Madrid, Spain}, 2019.
\newblock \doi{10.13009/EUCASS2019-949}.

\bibitem[{Ukai et~al.(2019)Ukai, Sakaki, Ishikawa, Sakaguchi, and
  Ishihara}]{ukai2019}
Ukai, S., Sakaki, K., Ishikawa, Y., Sakaguchi, H., and Ishihara, S.,
  \enquote{Component Tests of a {{LOX}}/Methane Full-Expander Cycle Rocket
  Engine: {{Injector}} and Regeneratively Cooled Combustion Chamber,}
  \emph{Proceedings of the 8th {{European Conference}} for {{Aeronautics}} and
  {{Space Sciences}}}, {Madrid, Spain}, 2019.

\bibitem[{Kajon et~al.(2019)Kajon, Liuzzi, Boffa, Rudnykh, DRIGO, Arione,
  Ierardo, and Sirbi}]{kajon2019}
Kajon, D., Liuzzi, D., Boffa, C., Rudnykh, M., DRIGO, D., Arione, L., Ierardo,
  N., and Sirbi, A., \enquote{Development of the Liquid Oxygen and Methane
  {{M10}} Rocket Engine for the {{Vega}}-{{E}} Upper Stage,} \emph{Proceedings
  of the 8th {{European Conference}} for {{Aeronautics}} and {{Space
  Sciences}}}, {Madrid, Spain}, 2019, p. 8 pages.
\newblock \doi{10.13009/EUCASS2019-315}.

\bibitem[{Traudt et~al.(2019)Traudt, Deeken, Oschwald, and
  Schlechtriem}]{traudt2019}
Traudt, T., Deeken, J.~C., Oschwald, M., and Schlechtriem, S., \enquote{Liquid
  {{Upper Stage Demonstrator Engine}} ({{LUMEN}}): {{Status}} of the
  {{Project}},} \emph{70th {{International Astronautical Congress}} ({{IAC}})},
  {Washington D.C., USA}, 2019.

\bibitem[{Soller et~al.(2014)Soller, Maeding, Kniesner, Preuss, Rackemann, and
  Blasi}]{soller2014}
Soller, S., Maeding, C., Kniesner, B., Preuss, A., Rackemann, N., and Blasi,
  R., \enquote{Characterisation of a {{LOX}}-{{LCH4 Gas Generator}},}
  \emph{Space {{Propulsion Conference}}}, {Cologne, Germany}, 2014, p.~10.

\bibitem[{B{\"o}rner et~al.(2018)B{\"o}rner, Manfletti, Hardi, Suslov, Kroupa,
  and Oschwald}]{borner2018}
B{\"o}rner, M., Manfletti, C., Hardi, J., Suslov, D., Kroupa, G., and Oschwald,
  M., \enquote{Laser Ignition of a Multi-Injector {{LOX}}/Methane Combustor,}
  \emph{CEAS Space Journal}, Vol.~10, No.~2, 2018, pp. 273--286.
\newblock \doi{10.1007/s12567-018-0196-6}.

\bibitem[{Haemisch et~al.(2019)Haemisch, Suslov, and Oschwald}]{haemisch2019}
Haemisch, J., Suslov, D., and Oschwald, M., \enquote{Experimental {{Study}} of
  {{Methane Heat Transfer Deterioration}} in a {{Subscale Combustion
  Chamber}},} \emph{Journal of Propulsion and Power}, Vol.~35, No.~4, 2019, pp.
  819--826.
\newblock \doi{10.2514/1.B37394}.

\bibitem[{{Waxenegger-Wilfing} et~al.(2020){Waxenegger-Wilfing}, Dresia,
  Deeken, and Oschwald}]{waxenegger-wilfing2020}
{Waxenegger-Wilfing}, G., Dresia, K., Deeken, J.~C., and Oschwald, M.,
  \enquote{Heat {{Transfer Prediction}} for {{Methane}} in {{Regenerative
  Cooling Channels}} with {{Neural Networks}},} \emph{Journal of Thermophysics
  and Heat Transfer}, Vol.~34, No.~2, 2020, pp. 347--357.
\newblock \doi{10.2514/1.T5865}.

\bibitem[{Balesdent(2012)}]{balesdent2012}
Balesdent, M., \enquote{Multidisciplinary {{Design Optimization}} of {{Launch
  Vehicles}},} Ph.D. thesis, \'Ecole centrale de Nantes, 2012.

\bibitem[{Wertz(2004)}]{wertz2004}
Wertz, J.~R., \enquote{Responsive {{Launch Vehicle Cost Model}},}
  \emph{Responsive {{Space Conference}}}, {Los Angeles,CA}, 2004, p.~13.

\bibitem[{Koelle(2013)}]{koelle2013}
Koelle, D.~E., \emph{Handbook of Cost Engineering for Space Transportation
  Systems with {{TRANSCOST}} 8.2 : Statistical-Analytical Model for Cost
  Estimation and Economical Optimization of Launch Vehicles}, {{TCS}}-{{TR}},
  {TransCostSystems}, 2013.

\bibitem[{Balesdent et~al.(2016)Balesdent, Brevault, Price, Defoort, Le~Riche,
  Kim, Haftka, and B{\'e}rend}]{balesdent2016}
Balesdent, M., Brevault, L., Price, N.~B., Defoort, S., Le~Riche, R., Kim,
  N.-H., Haftka, R.~T., and B{\'e}rend, N., \enquote{Advanced {{Space Vehicle
  Design Taking}} into {{Account Multidisciplinary Couplings}} and {{Mixed
  Epistemic}}/{{Aleatory Uncertainties}},} \emph{Space {{Engineering}}}, Vol.
  114, edited by G.~Fasano and J.~D. Pint{\'e}r, {Springer International
  Publishing}, {Cham}, 2016.
\newblock \doi{10.1007/978-3-319-41508-6_1}.

\bibitem[{Briese et~al.(2020)Briese, Acquatella~B, and Schnepper}]{briese2020}
Briese, L.~E., Acquatella~B, P., and Schnepper, K., \enquote{Multidisciplinary
  Modeling and Simulation Framework for Launch Vehicle System Dynamics and
  Control,} \emph{Acta Astronautica}, Vol. 170, 2020, pp. 652--664.
\newblock \doi{10.1016/j.actaastro.2019.08.022}.

\bibitem[{Vietze et~al.(2018)Vietze, Mundt, and Weiland}]{vietze2018}
Vietze, M., Mundt, C., and Weiland, S., \enquote{Cryogenic {{Launcher Stage
  Optimization Toolbox}},} \emph{Journal of Spacecraft and Rockets}, Vol.~55,
  No.~2, 2018, pp. 257--265.
\newblock \doi{10.2514/1.A33924}.

\bibitem[{Stappert et~al.(2019)Stappert, Wilken, Bussler, Sippel, Karl,
  Klevanski, Hantz, Briese, and Schnepper}]{stappert2019}
Stappert, S., Wilken, J., Bussler, L., Sippel, M., Karl, S., Klevanski, J.,
  Hantz, C., Briese, L.~E., and Schnepper, K., \enquote{European {{Next
  Reusable Ariane}} ({{ENTRAIN}}): {{A Multidisciplinary Study}} on a {{VTVL}}
  and a {{VTHL Booster Stage}},} \emph{Proceedings of the {{International
  Astronautical Congress}}, {{IAC}}}, {Washington DC}, 2019.

\bibitem[{Moroz et~al.(2019)Moroz, Burlaka, Barannik, Kochurov, and
  Maksiuta}]{moroz2019}
Moroz, L., Burlaka, M., Barannik, V., Kochurov, R., and Maksiuta, D.,
  \enquote{Liquid {{Rocket Propulsion Launcher Design System}} to {{Train
  AxSTREAM}}.{{AI}}. {{Reusability Aspects}},} 2019.

\bibitem[{Brevault et~al.(2020)Brevault, Balesdent, and Hebbal}]{brevault2020}
Brevault, L., Balesdent, M., and Hebbal, A., \enquote{Multi-{{Objective
  Multidisciplinary Design Optimization Approach}} for {{Partially Reusable
  Launch Vehicle Design}},} \emph{Journal of Spacecraft and Rockets}, Vol.~57,
  No.~2, 2020, pp. 373--390.
\newblock \doi{10.2514/1.A34601}.

\bibitem[{Christall(2017)}]{christall2017}
Christall, S., \enquote{Linking of {{Liquid Bipropellant Rocket Engine Design
  Methodology}} with {{Launch Vehicle Optimization}},} Master {{Thesis}},
  {University of W\"urzburg}, 2017.

\bibitem[{Jentzsch(2020)}]{jentzsch2020}
Jentzsch, S., \enquote{Optimization of a {{Reusable Launch Vehicle Using
  Genetic Algorithms}},} Master {{Thesis}}, {RWTH Aachen}, 2020.

\bibitem[{{Waxenegger-Wilfing} et~al.(2018){Waxenegger-Wilfing}, Dos
  Santos~Hahn, and Deeken}]{waxenegger-wilfing2018}
{Waxenegger-Wilfing}, G., Dos Santos~Hahn, R.~H., and Deeken, J.~C.,
  \enquote{Studies on {{Electric Pump}}-{{Fed Liquid Rocket Engines}} for
  {{Micro}}-{{Launcher}},} \emph{Proceedings - {{Space Propulsion}} 2018},
  {Seville, Spain}, 2018.

\bibitem[{Stappert et~al.(2018)Stappert, Wilken, Sippel, and
  Dumont}]{stappert2018}
Stappert, S., Wilken, J., Sippel, M., and Dumont, E., \enquote{{Assessment of a
  European Reusable VTVL Booster Stage},} \emph{{Space Propulsion 2018}},
  {Sevilla, Spanien}, 2018.

\bibitem[{McBride and Gordon(1996)}]{CEAmanual}
McBride, B., and Gordon, S., \enquote{Computer {{Program}} for {{Calculation}}
  of {{Complex Chemical Equilibrium Compositions}} and {{Applications}},}
  \emph{NASA Reference Publication 1311}, Vol. II. Users Manual and Program
  Description, 1996.

\bibitem[{Coley(1999)}]{coley1999}
Coley, D.~A., \emph{An {{Introduction}} to {{Genetic Algorithms}} for
  {{Scientists}} and {{Engineers}}}, {World Scientific}, 1999.
\newblock \doi{10.1142/3904}.

\bibitem[{B{\"a}ck(1996)}]{back1996}
B{\"a}ck, T., \emph{Evolutionary Algorithms in Theory and Practice: Evolution
  Strategies, Evolutionary Programming, Genetic Algorithms}, {Oxford University
  Press}, {New York}, 1996.

\bibitem[{Son et~al.(2014)Son, Ko, and Koo}]{son2014}
Son, M., Ko, S., and Koo, J., \enquote{Genetic Algorithm to Optimize the Design
  of Main Combustor and Gas Generator in Liquid Rocket Engines,} \emph{Journal
  of Thermal Science}, Vol.~23, No.~3, 2014, pp. 259--268.
\newblock \doi{10.1007/s11630-014-0704-8}.

\bibitem[{Marcus and Sedwick(2017)}]{marcus2017}
Marcus, M.~L., and Sedwick, R.~J., \enquote{Low {{Earth Orbit Debris Removal
  Technology Assessment Using Genetic Algorithms}},} \emph{Journal of
  Spacecraft and Rockets}, Vol.~54, No.~5, 2017, pp. 1110--1126.
\newblock \doi{10.2514/1.A33671}.

\bibitem[{Castellini and Lavagna(2012)}]{castellini2012a}
Castellini, F., and Lavagna, M.~R., \enquote{Comparative {{Analysis}} of
  {{Global Techniques}} for {{Performance}} and {{Design Optimization}} of
  {{Launchers}},} \emph{Journal of Spacecraft and Rockets}, Vol.~49, No.~2,
  2012, pp. 274--285.
\newblock \doi{10.2514/1.51749}.

\bibitem[{Pelamatti et~al.(2020)Pelamatti, Brevault, Balesdent, Talbi, and
  Guerin}]{pelamatti2020}
Pelamatti, J., Brevault, L., Balesdent, M., Talbi, E.-G., and Guerin, Y.,
  \enquote{Bayesian Optimization of Variable-Size Design Space Problems,}
  \emph{Optimization and Engineering}, 2020.
\newblock \doi{10.1007/s11081-020-09520-z}.

\bibitem[{Fortin et~al.(2012)Fortin, De~Rainville, Gardner, Parizeau, and
  {Christian Gagn\'e}}]{DEAP_JMLR2012}
Fortin, F.-A., De~Rainville, F.-M., Gardner, M.-A., Parizeau, M., and
  {Christian Gagn\'e}, \enquote{{{DEAP}}: {{Evolutionary}} Algorithms Made
  Easy,} \emph{Journal of Machine Learning Research}, Vol.~13, 2012, pp.
  2171--2175.

\end{thebibliography}

\end{document}